\documentclass[sigconf]{aamas} 

\usepackage{balance} 

\makeatletter
\gdef\@copyrightpermission{
  \begin{minipage}{0.2\columnwidth}
   \href{https://creativecommons.org/licenses/by/4.0/}{\includegraphics[width=0.90\textwidth]{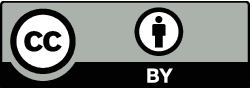}}
  \end{minipage}\hfill
  \begin{minipage}{0.8\columnwidth}
   \href{https://creativecommons.org/licenses/by/4.0/}{This work is licensed under a Creative Commons Attribution International 4.0 License.}
  \end{minipage}
  \vspace{5pt}
}
\makeatother

\setcopyright{ifaamas}
\acmConference[AAMAS '25]{Proc.\@ of the 24th International Conference
on Autonomous Agents and Multiagent Systems (AAMAS 2025)}{May 19 -- 23, 2025}
{Detroit, Michigan, USA}{Y.~Vorobeychik, S.~Das, A.~Nowé  (eds.)}
\copyrightyear{2025}
\acmYear{2025}
\acmDOI{}
\acmPrice{}
\acmISBN{}

\acmSubmissionID{{946}}

\usepackage[utf8]{inputenc} 
\usepackage[T1]{fontenc}    
\usepackage{hyperref}       
\usepackage{url}            
\usepackage{booktabs}       
\usepackage{amsfonts}       
\usepackage{nicefrac}       
\usepackage{microtype}      
\usepackage{xcolor}         

\usepackage{amsmath}
\usepackage{subcaption}
\usepackage{adjustbox}
 \usepackage{enumitem}
\usepackage{tabularx}
\usepackage{amsfonts}
\usepackage{cleveref}
\usepackage{tikz}
\usepackage{amsthm}
\usepackage{booktabs}
\usepackage{comment}
\usepackage{algorithm}
\usepackage[noend]{algpseudocode}

\newtheorem{theorem}{Theorem}
\newtheorem{lemma}{Lemma}
\newtheorem{definition}{Definition}
\usepackage{placeins}

\newcommand{\bm}[1]{\mathbf{#1}}
\newcommand{\reals}{\mathbb{R}}
\DeclareMathOperator*{\argmax}{arg\,max}

\definecolor{mydarkblue}{rgb}{0,0.2,0.7}

\title[AAMAS-2025 Formatting Instructions]{Multi-objective Reinforcement Learning with Nonlinear Preferences: Provable Approximation for Maximizing Expected Scalarized Return}

\author{Nianli Peng$^*$}
\affiliation{
  \institution{Harvard University}
  \city{Cambridge, MA}
  \country{USA}}
\email{nianli\_peng@g.harvard.edu}

\author{Muhang Tian$^*$}
\affiliation{
  \institution{Duke University}
  \city{Durham, NC}
  \country{USA}}
\email{muhang.tian@duke.edu}

\author{Brandon Fain}
\affiliation{
  \institution{Duke University}
  \city{Durham, NC}
  \country{USA}}
\email{btfain@cs.duke.edu}

\begin{abstract}
  We study multi-objective reinforcement learning with nonlinear preferences over trajectories. That is, we maximize the expected value of a nonlinear function over accumulated rewards (expected scalarized return or ESR) in a multi-objective Markov Decision Process (MOMDP). We derive an extended form of Bellman optimality for nonlinear optimization that explicitly considers time and current accumulated reward. Using this formulation, we describe an approximation algorithm for computing an approximately optimal non-stationary policy in pseudopolynomial time for smooth scalarization functions with a constant number of rewards. We prove the approximation analytically and demonstrate the algorithm experimentally, showing that there can be a substantial gap between the optimal policy computed by our algorithm and alternative baselines.
\end{abstract}

\keywords{Multi-objective Reinforcement Learning, Nonlinear Optimization, Algorithmic Fairness, Approximation Algorithms}

\newcommand{\BibTeX}{\rm B\kern-.05em{\sc i\kern-.025em b}\kern-.08em\TeX}

\begin{document}


\pagestyle{fancy}
\fancyhead{}


\maketitle 
\renewcommand{\thefootnote}{\fnsymbol{footnote}}
\footnotetext[1]{Equal contribution.}


\section{Introduction}
Markov Decision Processes (MDPs) model goal-driven interaction with a stochastic environment, typically aiming to maximize a scalar-valued reward per time-step through a learned policy. Equivalently, this formulation asks the agent to maximize the expected value of a \emph{linear} function of total reward. This problem can be solved with provable approximation guarantees in polynomial time with respect to the size of the MDP \cite{kearns02,brafman03,auer08,azar17,agarwal21}.

We extend this framework to optimize a \emph{nonlinear} function of vector-valued rewards in multi-objective MDPs (MOMDPs), aiming to maximize expected \emph{welfare} $\mathbb{E}[W(\mathbf{r})]$ where $\mathbf{r}$ is a total reward vector for $d$ objectives. We note that this function is also called the \textit{utility} or the \textit{scalarization} function within the multi-objective optimization literature \cite{hayes23}. Unlike the deep neural networks commonly used as function approximators for large state spaces, our nonlinearity lies entirely in the objective function.

\paragraph{Motivation.} Nonlinear welfare functions capture richer preferences for agents, such as fairness or risk attitudes. For example, the Nash Social Welfare function $\left(W_{\text{Nash}}(\mathbf{r}) = \left(\prod_{i} r_i\right)^{1/d}\right)$ reflects a desire for fairness or balance across objectives and diminishing marginal returns \cite{caragiannis19}. Even single-objective decision theory uses nonlinear utility functions to model risk preferences, like risk aversion with the Von Neumann-Morgenstern utility function \cite{VNM47}. 


Consider a minimal example with an autonomous taxi robot, Robbie, serving rides in neighborhoods $A$ and $B$, as diagrammed in Figure~\ref{fig:example}. Each ride yields a reward in its respective dimension.

\begin{figure}[ht]
\centering
\begin{tikzpicture}[thick, ->, inner sep=8pt]
\node (A) at (-2, 0) {$A$}; \draw (-2, 0) circle (12pt);
\node (B) at (2, 0) {$B$}; \draw (2, 0) circle (12pt);

\draw (A) to [loop left] (A); \node at (-2.8, 0.4) {$(1, 0)$};
\draw (B) to [loop right] (B); \node at (2.8, 0.4) {$(0, 1)$};
\draw (A) to [out=340, in=200] (B); \node at (0, 0) {$(0, 0)$};
\draw (B) to [out=160, in=20] (A);

\end{tikzpicture}
\caption{Taxi Optimization Example}
\label{fig:example}
\end{figure}
Suppose Robbie starts in neighborhood $A$ and has $t=3$ time intervals remaining before recharging. With no discounting, the \textit{Pareto Frontier} of maximal (that is, undominated) policies can achieve cumulative reward vectors of $(3, 0)$, $(1, 1)$, or $(0, 2)$ by serving $A$ alone, $A$ and $B$, or $B$ alone respectively. If we want Robbie to prefer the second more balanced or ``fair'' option of serving one ride in each of the two neighborhoods then Robbie must have \textit{nonlinear} preferences. That is, for any choice of weights on the first and second objective, the simple weighted average would prefer outcomes (3, 0) or (0, 2). However, the second option would maximize the Nash Social Welfare of cumulative reward, for example.



Nonlinear preferences complicate policy computation: Bellman optimality fails, and stationary policies may be suboptimal. Intuitively, a learning agent with fairness-oriented preferences to balance objectives should behave differently, even in the same state, depending on which dimension of reward is ``worse off.'' In the Figure~\ref{fig:example} example, one policy to achieve the balanced objective $(1, 1)$ is to complete a ride in $A$, then travel from $A$ to $B$, and finally to complete a ride in $B$ -- note that this is not stationary with respect to the environment states.


\paragraph{Contributions.} In this work, we ask whether it is possible to describe an approximation algorithm (that is, with provable guarantees to approximate the welfare optimal policy) for MOMDPs with nonlinear preferences that has polynomial dependence on the number of states and actions, as is the case for linear preferences or scalar MDPs. To the best of our knowledge, ours is the first work to give provable guarantees for this problem, compared to other work that focuses on empirical evaluation of various neural network architectures.

We show this \textit{is} possible for smooth preferences and a constant number of dimensions of reward. To accomplish this, we (i) derive an extended form of Bellman optimality (which may be of independent interest) that characterizes optimal policies for nonlinear preferences over multiple objectives, (ii) describe an algorithm for computing approximately optimal non-stationary policies, (iii) prove the worst-case approximation properties of our algorithm, and (iv) demonstrate empirically that our algorithm can be used in large state spaces to find policies that significantly outperform other baselines.

\section{Related Work}
Most reinforcement learning algorithms focus on a single scalar-valued objective and maximizing total expected reward \cite{sutton18}. Classic results on provable approximation and runtime guarantees for reinforcement learning include the E3 algorithm \cite{kearns02}. This result showed that general MDPs could be solved to near-optimality efficiently, meaning in time bounded by a polynomial in the size of the MDP (number of states and actions) and the horizon time. Subsequent results refined the achievable bounds \cite{brafman03,auer08,azar17}. We extend these results to the multi-objective case with nonlinear preferences.

Multi-objective reinforcement learning optimizes multiple objectives at once. So-called single-policy methods use a scalarization function to reduce the problem to scalar optimization for a single policy, and we follow this line of research. The simplest form is linear scalarization, applying a weighted sum to the Q vector \cite{moffaert2013hypervolume, liu2014multiobjective, abels19, yang19, alegre23}.

A more general problem is to optimize the expected value of a potentially nonlinear function of the total reward, which may be vector-valued in a multi-objective optimization context. We refer to such a function as a welfare function \cite{barman2023fairness, fan23, siddique20}, which is also commonly referred to as a utility or scalarization function \cite{hayes23, agarwal22}. Recent works have explored nonlinear objectives; however, to our knowledge, ours is the first to provide an approximation algorithm with provable guarantees (on the approximation factor), for the expected welfare, leveraging a characterization of recursive optimality in this setting. Several other studies focus on algorithms that demonstrate desirable convergence properties and strong empirical performance by conditioning function approximators on accumulated reward, but without offering approximation guarantees \cite{siddique20, fan23, cai23, reymond23}. Complementary to our approach, \cite{reymond2022pareto} uses Pareto Conditioned Networks to learn policies for Pareto-optimal solutions by conditioning policies on a preference vector. \cite{lin2024offline} presents an offline adaptation framework that employs demonstrations to implicitly learn preferences and safety constraints, aligning policies with inferred preferences rather than providing theoretical guarantees.

\cite{agarwal22} describe another model-based algorithm that can compute approximately optimal policies for a general class of monotone and Lipschitz welfare functions, but rather than maximizing the expected welfare, they maximize the welfare of expected rewards (note the two are not equal for nonlinear welfare functions). Other works have formulated fairness in different ways or settings. For example, \cite{jabbari2017fairness} defines an analogue of envy freeness and \cite{pmlr-v206-deng23a} studies a per-time-step fairness guarantee. \cite{barman2023fairness} studies welfare maximization in multi-armed bandit problems. \cite{roepke2023distributional} explores the concept of distributional multi-objective decision making for managing uncertainty in multi-objective environments.

Risk-sensitive RL approaches address scalar objectives by incorporating risk measures to minimize regret or control reward variance over accumulated rewards \cite{bauerle2011markov, bellemare2023distributional, bastani2022regret}. While these approaches offer valuable tools for managing reward variability, their guarantees are primarily in terms of regret minimization or achieving bounded variance. In contrast, our work provides stronger theoretical assurances in terms of the approximation ratio on the expected welfare for multi-objective optimization. This difference in focus underscores the robustness of our method, which provides guarantees that extend beyond the risk-sensitive regime to cover complex, multi-dimensional utility functions \cite{fan23, brafman03, azar17}. Such problems remain computationally significant even in a deterministic environment where the notion of risk may not apply.

Lastly, while our single-agent setup with multiple objectives shares some aspects with multi-agent reinforcement learning (MARL), the objectives differ significantly. Much of the MARL literature has focused on cooperative reward settings, often using value-decomposition techniques like VDN and QMIX \cite{sunehag2017value, rashid2018qmix} or actor-critic frameworks to align agent objectives under a centralized training and decentralized execution paradigm \cite{foerster2018counterfactual}. In contrast, our work parallels the more general Markov game setting, where each agent has a unique reward function and studies nonlinear objectives that require computational methods beyond linear welfare functions often assumed in MARL. While MARL research frequently uses linear summations of agent rewards, we demonstrate approximation guarantees for optimizing general, non-linear functions of multiple reward vectors, a distinct contribution in a single-agent setting \cite{jabbari2017fairness, agarwal22, VNM47}.

\section{Preliminaries}
\label{section:preliminaries}
A finite Multi-objective Markov Decision Process (MOMDP) consists of a finite set $\mathcal{S}$ of states, a starting state $s_1\in\mathcal{S}$,\footnote{In general we may have a distribution over starting states; we assume a single starting state for ease of exposition.} a finite set $\mathcal{A}$ of actions, and probabilities $Pr(s'|s,a) \in [0, 1]$ that determine the probability of transitioning to state $s'$ from state $s$ after taking action $a$. Probabilities are normalized so that $\sum_{s'} Pr(s'|s,a) = 1$.

We have a finite vector-valued reward function $\mathbf{R}(s,a): \mathcal{S} \times \mathcal{A} \rightarrow [0,1]^d$. Each of the $d$ dimensions of the reward vector corresponds to one of the multiple objectives that are to be maximized. At each time step $t$, the agent observes state $s_t \in \mathcal{S}$, takes action $a_t \in \mathcal{A}(s_t)$, and receives a reward vector $\bm{R}(s_t, a_t) \in [0,1]^d$. The environment, in turn, transitions to $s_{t+1}$ with probability $Pr(s_{t+1}|a_t,s_t)$.




To make the optimization objective well-posed in MOMDPs with vector-valued rewards, we must specify a \textit{scalarization} function \cite{hayes23} which we denote as $W: \reals^d \to \reals$.
For fair multi-objective reinforcement learning, we think of each of the $d$ dimensions of the reward vector as corresponding to distinct users. The scalarization function can thus be thought of as a \textit{welfare} function  over the users, and the learning agent is a welfare maximizer. Even when $d=1$, a nonlinear function $W$ can be a Von Neumann-Morgenstern utility function \cite{VNM47} that expresses the risk-attitudes of the learning agent, with strictly concave functions expressing risk aversion.

\paragraph{Assumptions.} Here we clarify some preliminary assumptions we make about the reward function $\mathbf{R}(s,a): \mathcal{S} \times \mathcal{A} \rightarrow [0,1]^d$ and the welfare function $W: \mathbb{R}^d \to \mathbb{R}$. The restriction to $[0,1]$ is simply a normalization for ease of exposition; the substantial assumption is that the rewards are finite and bounded. Note from the writing we assume the reward function is deterministic: While it suffices for linear optimization to simply learn the mean of random reward distributions, this does not hold when optimizing the expected value of a nonlinear function. Nevertheless, the environment itself may still be stochastic, as the state transitions may still be random.

We also assume that $W$ is \textbf{smooth}. For convenience of analysis, we will assume $W$ is \textbf{uniformly continuous} on the L1-norm (other parameterizations such as the stronger Lipshitz continuity or using the L2-norm are also possible). For all $\epsilon > 0$, there exists $\delta_\epsilon$ such that for all $\mathbf{x}, \mathbf{y} \geq 0$, 
$$|W(\mathbf{x}) - W(\mathbf{y})| < \epsilon \quad \text{if} \quad \|\mathbf{x} - \mathbf{y}\|_1 < \delta_\epsilon.$$
The smoothness assumption seems necessary to give worst-case approximation guarantees as otherwise arbitrarily small changes in accumulated reward could have arbitrarily large differences in welfare. However, we note that this is still significantly more general than linear scalarization which is implicitly smooth. Practically speaking, our algorithms can be run regardless of the assumed level of smoothness; a particular smoothness is necessary just for the worst-case analysis.

\section{Modeling Optimality}
\label{section:model}
In this section, we expand the classic model of reinforcement learning to optimize the expected value of a nonlinear function of (possibly) multiple dimensions of reward. We begin with the notion of a trajectory of state-action pairs.

\begin{definition}
    Let $M$ be an MOMDP. A length $T$ \textit{trajectory} in $M$ is a tuple $\tau$ of $T$ state-action pairs $$(s_1, a_1), (s_2, a_2), \dots ,(s_T, a_T).$$
    For $1 \leq k \leq k' \leq T$, let $\tau_{k:k'}$ be the sub-trajectory consisting of pairs $(s_k, a_k), \dots, (s_{k'}, a_{k'})$. Let $\tau_{0;0}$ denote the empty trajectory.
\end{definition}

For a \textit{discount factor} $\gamma$, we calculate the total discounted reward of a trajectory. Note that this is a vector in general.

\begin{definition}
For length $T$ trajectory $\tau$ and discount factor $0 \le \gamma \leq 1$, the \textit{total discounted reward} along $\tau$ is the vector
$$
\bm{R}(\tau, \gamma) = \sum_{t=1}^{T} \gamma^{t-1} \bm{R}(s_t, a_t).
$$
For ease of exposition we will frequently leave $\gamma$ implicit from context and simply write $\bm{R}(\tau)$. \footnote{$\gamma < 1$ is necessary for the infinite horizon setting. In the experiments with a finite-horizon task we use $\gamma = 1$ for simplicity.}
\end{definition}

A \textit{policy} is a function $\pi(a \mid \tau, s) \in [0, 1]$ mapping past trajectories and current states to probability distributions over actions, that is, $\sum_{a} \pi(a \mid \tau, s) = 1$ for all $\tau$ and $s$. A \textit{stationary policy} is the special case of a policy that depends only on the current state: $\pi(a \mid s)$.

\begin{definition}
    The probability that a $T$-trajectory $\tau$ is traversed in an MOMDP $M$ upon starting in state $s_1$ and executing policy $\pi$ is
    \begin{equation*}
        Pr^\pi[\tau] = \pi (a_1 \mid \tau_{0:0}, s_1 ) \times \prod_{t=2}^T \pi\left(a_t ~\bigg|~ \tau_{1:t-1}, s_t \right) Pr(s_{t}|s_{t-1}, a_{t-1}).
    \end{equation*}
    
\end{definition}

\paragraph{Problem Formulation.}


Given a policy, a finite time-horizon $T$, and a starting state $s_1$ we can calculate the expected welfare of total discounted reward along a trajectory as follows. Our goal is to maximize this quantity. That is, we want to compute a policy that maximizes the expected $T$-step discounted welfare. 

\begin{definition}
    For a policy $\pi$ and a start state $s$, the \textit{expected $T$-step discounted welfare} is 
    $$
    \mathbb{E}_{\tau \sim \pi} \bigl[W \bigl(\bm{R}(\tau) \bigr) \bigr] = \sum_\tau Pr^\pi[\tau]W(\mathbf{R}(\tau))
    $$
    where the expectation is taken over all length $T$ trajectories beginning at $s$.
\end{definition}

Note that this objective is not equal to $W \left( \mathbb{E}_{\tau \sim \pi} \bigl[ \bm{R}(\tau) \bigr] \right)$, which others have studied \cite{agarwal22,siddique20}, for a nonlinear $W$. The former (our objective) is also known as \textit{expected scalarized return} (ESR) whereas the latter is also known as \textit{scalarized expected return} (SER) \cite{hayes23}. While SER makes sense for a repeated decision-making problem, it does not optimize for expected welfare for any particular trajectory. For concave $W$, ESR $\leq$ SER by Jensen's inequality. However, an algorithm for approximating SER does not provide any guarantee for approximating ESR. For example, a policy can be optimal on SER but achieve 0 ESR if it achieves high reward on one or the other of two objectives but never both in the same episode. 

\paragraph{Form of Optimal Policy and Value Functions.}
The optimal policy for this finite time horizon setting is a function also of the number of time steps remaining. We write such a policy as $\pi(a \mid s, \tau, t)$ where $\tau$ is a trajectory (the history), $s$ is the current state, and $t$ is the number of time steps remaining, $i.e.$ $t = T - |\tau|$. 

We can similarly write the extended value function of a policy $\pi$. We write $\tau$ as the history or trajectory prior to some current state $s$ and $\tau'$ as the future, the remaining $t$ steps determined by the policy $\pi$ and the environmental transitions. 

\begin{definition}
\label{definition: Vstar}
    The value of a policy $\pi$ beginning at state $s$ after history $\tau$ and with $t$ more actions is 
    $$
    V^{\pi}(s, \tau, t) = \mathbb{E}_{\tau' \sim \pi} \left[ W(\mathbf{R}(\tau) + \gamma^{T-t} \mathbf{R}(\tau'))\right]
    $$
    where the expectation is taken over all length $t$ trajectories $\tau'$ beginning at $s$. The optimal value function is 
    $$
    V^{*}(s, \tau, t) = \max_{\pi} V^{\pi}(s, \tau, t)
    $$ and the optimal policy is 
    $$
    \pi^* \in \argmax_{\pi} V^{\pi}(s, \tau, t).
    $$
\end{definition}

Note that because $W$ is nonlinear, the value function $V^{\pi}$ cannot be decomposed in the same way as in the traditional Bellman equations. Before proceeding we want to provide some intuition for this point. The same reasoning helps to explain why stationary policies are not generally optimal. 


Suppose we are optimizing the product of the reward between two objectives (i.e., the welfare function is the product or geometric mean), and at some state $s$ with some prior history $\tau$ we can choose between two policies $\pi_1$ or $\pi_2$. Suppose that the future discounted reward vector under $\pi_1$ is $(1, 1)$, whereas it is $(10, 0)$ under $\pi_2$. So $\pi_1$ has greater expected future welfare and traditional Bellman optimality would suggest we should choose $\pi_1$. However, if $\mathbf{R}(\tau) = (0, 10)$, we would actually be better off in terms of total welfare choosing $\pi_2$. In other words, both past and future reward are relevant when optimizing for the expected value of a nonlinear welfare function.

We develop an extended form of Bellman optimality capturing this intuition by showing that the optimal value function can be written as a function of current state $s$, accumulated discounted reward $\mathbf{R}(\tau)$, and number of timesteps remaining in the task $t$. The proof is included in the appendix. At a high level as a sketch, the argument proceeds inductively on $t$ where the base case follows by definition and the inductive step hinges on showing that the the expectation over future trajectories can be decomposed into an expectation over successor states and subsequent trajectories despite the nonlinear $W$.

\begin{lemma}
\label{lemma: sufficiency}
    Let $\mathcal{V}(s, \mathbf{R}(\tau), 0) = W(\mathbf{R}(\tau))$ for all states $s$ and trajectories $\tau$. For every state $s$, history $\tau$, and $t > 0$ time steps remaining, let
    $$
    \mathcal{V}(s, \mathbf{R}(\tau), t) = \max_a \mathbb{E}_{s'} \left[ \mathcal{V} (s', \mathbf{R}(\tau) + \gamma^{T-t}\mathbf{R}(s,a), t-1) \right].
    $$
    Then $V^*(s, \tau, t) = \mathcal{V}(s, \mathbf{R}(\tau), t)$.
\end{lemma}


By Lemma \ref{lemma: sufficiency}, we can parameterize the optimal value function by the current state $s$, accumulated reward $\mathbf{R_{acc}} \in \reals^d$, and number of timesteps remaining $t$. We will use this formulation of $V^*$ in the remainder of the paper.

\begin{definition}[Recursive Formulation of $V^*$] Let $\mathbf{R_{acc}} = \mathbf{R}(\tau)$ be the vector of accumulated reward along a history prior to some state $s$. Let $V^*(s, \mathbf{R_{acc}}, 0) = W(\mathbf{R_{acc}})$ for all $s$ and $\mathbf{R_{acc}}$. For $t > 0$,
$$V^*(s, \mathbf{R}_{acc}, t) = \max_a \sum_{s'}Pr(s'|s, a) \cdot V^*(s', \mathbf{R}_{acc} + \gamma^{T-t}\mathbf{R}(s,a), t-1).$$ \label{definition:recvstar}
\end{definition}

\paragraph{Horizon Time.} Note that an approximation algorithm for this discounted finite time horizon problem can also be used as an approximation algorithm for the discounted infinite time horizon problem. Informally, discounting by $\gamma$ with bounded maximum reward implies that the first $T \approx 1/(1 - \gamma)$ steps dominate overall returns. We defer the precise formulation of the lower bound on the horizon time and its proof in the appendix (Lemma \ref{lemma:horizon}).

\paragraph{Necessity of Conditioning on Remaining Timesteps \( t \)}

We illustrate the necessity of conditioning the optimal value function on the remaining timesteps $t$ using a counterexample in the appendix.

\section{Computing Optimal Policies}
\label{section:algorithm}
Our overall algorithm is \textsc{Reward-Aware Explore or Exploit} (or \textsc{RAEE} for short) to compute an approximately optimal policy, inspired by the classical E3 algorithm \cite{kearns02}. At a high level, the algorithm explores to learn a model of the environment, periodically pausing to recompute an approximately optimal policy on subset of the environment that has been thoroughly explored. We call this optimization subroutine \textsc{Reward-Aware Value Iteration} (or \textsc{RAVI} for short). In both cases, reward-aware refers to the fact that the algorithms compute non-stationary policies in the sense that optimal behavior depends on currently accumulated vector-valued reward, in addition to the current state.

Of the two, the model-based optimization subroutine \textsc{RAVI} is the more significant. The integrated model-learning algorithm \textsc{RAEE} largely follows from prior work, given access to the \textsc{RAVI} subroutine. For this reason, we focus in this section on \textsc{RAVI}, and defer a more complete discussion and analysis of \textsc{RAEE} to Section~\ref{section:remove} and the appendix.

A naive algorithm for computing a non-stationary policy would need to consider all possible prior trajectories for each decision point, leading to a runtime complexity containing the term $|\mathcal{S}|^T$, exponential in the size of the state space and time horizon. Instead, for a smooth welfare function on a constant number of objectives, our algorithm will avoid any exponential dependence on $|\mathcal{S}|$. 

The algorithm is derived from the recursive definition of the optimal multi-objective value function $V^*$ in Definition~\ref{definition:recvstar}, justified by Lemma \ref{lemma: sufficiency}, parameterized by the accumulated discounted reward vector $\mathbf{R_{acc}}$ instead of the prior history. Note that even if the rewards were integers, $\mathbf{R_{acc}}$ might not be due to discounting. We must therefore introduce a discretization that maps accumulated reward vectors to points on a lattice, parameterized by some $\alpha \in (0, 1)$ where a smaller $\alpha$ leads to a finer discretization but increases the runtime. 


\begin{definition}
For a given discretization precision parameter $\alpha \in \mathbb{R}^+$, define $f_{\alpha}: \reals^d \to (\alpha\mathbb{Z})^d$ by
$$
f_{\alpha}(\mathbf{R}) = \left(  \left\lfloor \frac{R_1}{\alpha} \right\rfloor \cdot \alpha,   \left\lfloor \frac{R_2}{\alpha} \right\rfloor \cdot \alpha, \cdots, \left\lfloor \frac{R_d}{\alpha} \right\rfloor \cdot \alpha \right).
$$
\end{definition}

In other words, $f_{\alpha}$ maps any $d$-dimensional vector to the largest vector in $(\alpha \mathbb{Z})^d$ that is less than or equal to the input vector, effectively rounding each component \textit{down} to the nearest multiple of $\alpha$. We now describe the algorithm, which at a high level computes the dynamic program of approximately welfare-optimal value functions conditioned on discretized accumulated reward vectors.

\textbf{Remark.} Since we model the per-step reward as normalized to at most 1, we describe $\alpha$ as lying within $(0,1)$. However, the algorithm itself is well-defined for larger values of alpha (coarser than the per-step reward) and during implementation and experiments, we consider such larger $\alpha$, beyond what might give worst-case guarantees but still observing strong empirical performance.

\begin{algorithm}
\caption{\textsc{Reward-Aware Value Iteration (RAVI)}\label{alg:cap}}
\begin{algorithmic}[1]
\State \textbf{Parameters:} Discretization precision $\alpha \in (0, 1)$, Discount factor $\gamma \in [0, 1)$, Reward dimension $d$, finite time horizon $T$, welfare function $W$, discretization function $f_{\alpha}$.
\State \textbf{Require:} Normalize $\mathbf{R}(s,a) \in [0,1]^d$ for all $s \in \mathcal{S}, a \in \mathcal{A}$. 
\State \textbf{Initialize:} $V(s, \mathbf{R_{acc}}, 0) = W(\mathbf{R_{acc}})$ for all $\mathbf{R_{acc}} \in \{0, \mathbf{\alpha}, 2\mathbf{\alpha}, \dots, \lceil T/\alpha\rceil \cdot \alpha\}^d$, $s \in \mathcal{S}$.
\For{t = 1 to $T$}
\For{$\mathbf{R_{acc}} \in \{0, \mathbf{\alpha}, 2\mathbf{\alpha}, \dots, \lceil (T - t) / \alpha \rceil \cdot \alpha \}^d$}
\For{all $s \in \mathcal{S}$}
\State \begin{equation*}
    V\left(s,\mathbf{R_{acc}}, t\right) \gets \max_a \sum_{s'}Pr\left(s'|s, a\right) V\left(s', \varphi(s,\mathbf{R}_{acc}, a), t-1\right)
\end{equation*}
\State \begin{equation*}
    \pi\left(s,\mathbf{R_{acc}}, t\right) \gets \argmax_a \sum_{s'}Pr\left(s'|s, a\right) V\left(s', \varphi(s, \mathbf{R}_{acc}, a), t-1\right)
\end{equation*}
\State where $\varphi(s,\mathbf{R}_{acc},a) := f_{\alpha} \left(\mathbf{R}_{acc} + \gamma^{T-t}\mathbf{R}(s,a)\right)$
\EndFor
\EndFor
\EndFor
\end{algorithmic}
\end{algorithm}
The asymptotic runtime complexity is $O\left(|\mathcal{S}|^2 |\mathcal{A}| (T/\alpha)^d \right)$. However, observe that the resulting algorithm is extremely parallelizable: Given solutions to subproblems at $t-1$, all subproblems for $t$ can in principle be computed in parallel. A parallel implementation of \textsc{RAVI} can leverage GPU compute to handle the extensive calculations involved in multi-objective value iteration. Each thread computes the value updates and policy decisions for a specific state and accumulated reward combination, which allows for massive parallelism. In practice, we observed an empirical speedup of approximately 560 times on a single NVIDIA A100 compared to the CPU implementation on an AMD Ryzen 9 7950x 16-core processor for our experimental settings. This drastic improvement in runtime efficiency makes it feasible to run \textsc{RAVI} on larger and more complex environments, where the computational demands would otherwise be prohibitive.

It remains to see how small $\alpha$ needs to be, which will dictate the final runtime complexity. We first analyze the correctness of the algorithm and then return to the setting of $\alpha$ for a given smoothness of $W$ to conclude the runtime analysis.

\label{section:analysis}

We begin analyzing the approximation of \textsc{RAVI} by showing an important structural property: the optimal value function will be smooth as long as the welfare function is smooth. The proof is included in the appendix.

\begin{lemma}[Uniform continuity of multi-objective value function]
\label{lemma:lipschitz}
Let the welfare function $W: \mathbb{R}^d \to \mathbb{R}$ be uniformly continuous. Fix $s \in \mathcal{S}$ and $t \in \{0,1,\dots, T\}$, then for all $\epsilon > 0$, there exists $\delta_\epsilon > 0$ such that
$$
\bigg|V^*(s, \mathbf{R_1}, t) - V^*(s, \mathbf{R_2}, t) \bigg| < \epsilon \quad \text{if} \quad \|\mathbf{R_1} - \mathbf{R_2}\| < \delta_\epsilon.
$$
\end{lemma}

We now present the approximation guarantee of \textsc{RAVI}, that it achieves an additive error that scales with the smoothness of $W$ and the number of remaining time steps. The proof of this lemma is deferred to the appendix. 
\begin{lemma}[Approximation Error of \textsc{RAVI}]
\label{lemma:error}
For uniformly continuous welfare function $W$, for all $\epsilon > 0$, there exists $\alpha_\epsilon$ such that 
    $$
    V(s, \mathbf{R_{acc}}, t) \ge V^*(s, \mathbf{R_{acc}}, t) - t\epsilon
    $$
    \quad $\forall s \in \mathcal{S}, \mathbf{R_{acc}} \in \reals^d, t \in \{0, 1, \dots, T\}$, where $V(s, \mathbf{R_{acc}}, t)$ is computed by Algorithm~\ref{alg:cap} using $\alpha_\epsilon$.
\end{lemma}

While we can use any setting of $\alpha$ empirically, this shows that for an approximation guarantee we should set the discretization parameter to $\alpha = \delta_{T\epsilon}/d$ where $\delta_{T\epsilon}$ from the smoothness of the welfare function is sufficient to drive its bound to $\epsilon$, that is, it should be $\alpha = \frac{\delta_{\epsilon}}{T \cdot d}.$

We thus arrive at the ultimate statement of the approximation and runtime of \textsc{RAVI}. The proof follows directly from Lemma~\ref{lemma:error} and the setting of $\alpha$.
\begin{theorem}[Optimality Guarantee of \textsc{RAVI}]
\label{theorem:value iteration}
For a given $\epsilon$ and welfare function $W$ that is $\delta_{\epsilon}$ uniformly continuous, \textsc{RAVI} with $\alpha = \frac{\delta_{\epsilon}}{T \cdot d}$ computes a policy $\hat{\pi}$, such that $V^{\hat{\pi}} (s, 0, T) \ge V^*(s, 0, T) -\epsilon$ in $O\left(|\mathcal{S}|^2 |\mathcal{A}| (d \cdot T^2/\delta_{\epsilon})^d \right)$ time. 
\end{theorem}

A concrete example of a particular welfare function, the setting of relevant parameters, and a derivation of a simplified runtime statement may help to clarify Theorem~\ref{theorem:value iteration}. Consider the smoothed \textit{proportional fairness} objective: $W_{\text{SPF}}(\mathbf{x}) = \sum_{i=1}^d \ln(x_i + 1)$, a smoothed log-transform of the Nash Social Welfare (or geometric mean) with better numerical stability. 

By taking the gradient, we can see that $W_{\text{SPF}}(\mathbf{x})$ is $d$-Lipschitz on $\mathbf{x} \ge 0$, so we may pick $\delta_\epsilon = \epsilon / d$ to satisfy the uniform continuity requirement in Theorem~\ref{theorem:value iteration}. \footnote{Uniform continuity is a weaker modeling assumption than $L$-Lipshitz continuity for constant $L$. Note that for an $L$-Lipshitz function, the correct value of $\delta_{\epsilon}$ is just $\epsilon / L$, where $\epsilon$ is the desired approximation factor of the algorithm.}



Plugging $\delta_\epsilon$ into the runtime and recalling that $\alpha_{T\epsilon} = \tfrac{\delta_{T\epsilon}}{d}$ and $\delta_{T\epsilon} = \delta_\epsilon / T$ up to constant factors, we get
\begin{equation*}
O\left(|\mathcal{S}|^2 |\mathcal{A}| (T/\alpha_{T\epsilon})^d \right)
= O\left(|\mathcal{S}|^2 |\mathcal{A}| \left(\frac{T^2 \cdot d^2}{\epsilon} \right)^d \right).
\end{equation*}
To further simplify, if one takes the number of actions $|\mathcal{A}|$, discount factor $\gamma$, and dimension $d$ to be constants, then the asymptotic dependence of the runtime (in our running example) is
$$O\left( |\mathcal{S}|^2 \left( \frac{1}{\epsilon} \log^2(1/\epsilon)\right)^d \right).$$

This dependence is significantly better than a naive brute-force approach, which scales at least as $O\bigl(|\mathcal{S}|^T\bigr)$.

Intuitively, the key savings arise from discretizing the reward space (with granularity $\delta_{\epsilon}$) rather than enumerating all possible trajectories. This discretization is guided by the smoothness assumption on $W_{\text{SPF}}$.

\section{Experiments}
\label{section:experiment}
\begin{figure}
    \begin{subfigure}{0.5\linewidth}
        \centering
        \includegraphics[width=\linewidth]{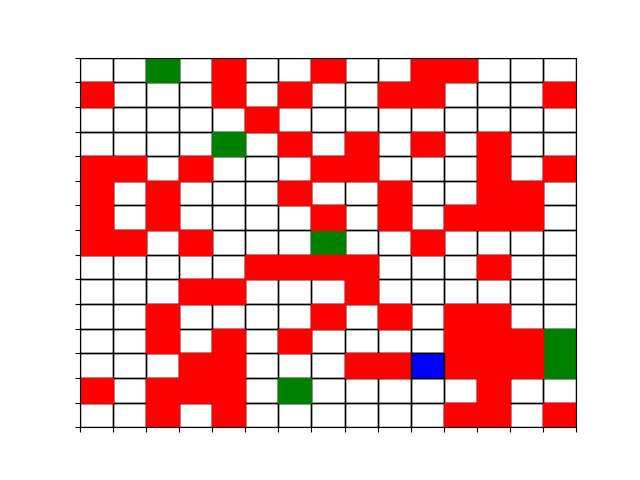}
        \caption*{Scavenger}
    \end{subfigure}%
    \begin{subfigure}{0.35\linewidth}
        \centering
        \includegraphics[width=\linewidth]{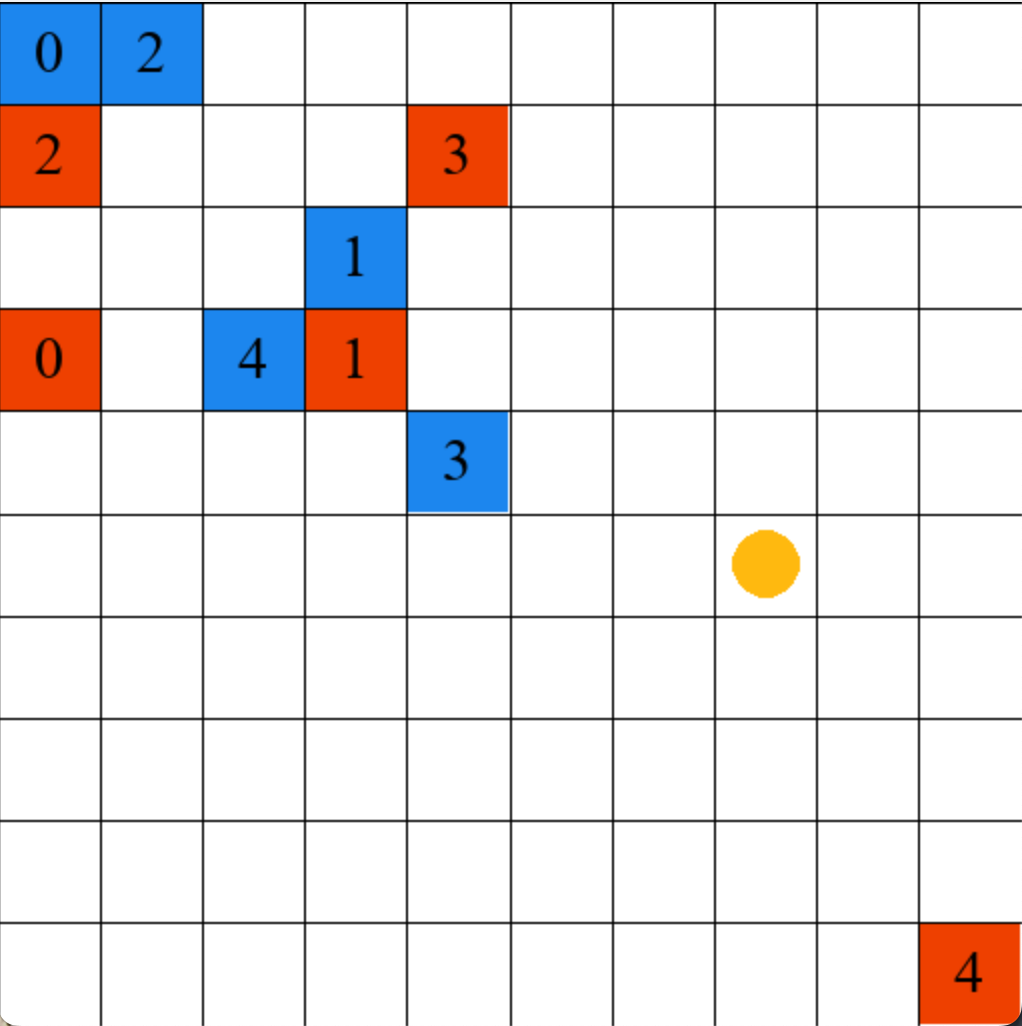}
        \caption*{Taxi, $d=5$}
    \end{subfigure}%
    \caption{Visualization of the Taxi and Scavenger environments.}
    \label{env vis}
\end{figure}
We test \textsc{RAVI} on two distinct interpretations of multi-objective reinforcement learning: (1) the fairness interpretation, where the agent tries to maximize rewards across all dimensions. (2) the objective interpretation, where the agent tries to maximize one while minimizing the other. In both scenarios, we show that \textsc{RAVI} can discover more optimal policies than other baselines for nonlinear multi-objective optimization. This holds even for coarser settings of $\alpha$ in the algorithm than would be necessary for strong theoretical worst-case approximation guarantees.

\subsection{Simulation Environments}
\textbf{Taxi}: We use the taxi multi-objective simulation environment considered in \cite{fan23} for testing nonlinear ESR maximization.
In this grid-world environment, the agent is a taxi driver whose goal is to deliver passengers. There are multiple queues of passengers, each having a given pickup and drop-off location. Each queue has a different objective or dimension of the reward. The taxi can only take one passenger at a time and wants to balance trips for passengers in the different queues. This environment models the fairness interpretation.

\textbf{Scavenger}: Inspired by the Resource Gathering environment \cite{10.1145/1390156.1390162}, the scavenger hunt environment is a grid-world simulation where the agent must collect resources while avoiding enemies scattered across the grid. The state representation includes the agent's position and the status of the resources (collected or not). The reward function is vector-valued, where the first component indicates the number of resources collected, and the second component indicates the damage taken by encountering enemies. This environment is the objective interpretation.

\subsection{Baseline Algorithms}

\textbf{Linearly Scalarized Policy} (LinScal) \cite{van2013scalarized}: A relatively straightforward technique for multi-objective RL optimization is to apply the linear combination on the Q-values for each objective.
Given weights $\mathbf{w} \in \mathbb{R}^d$, $\sum_{i}^{d} w_{i} = 1$, the scalarized objective is $SQ(s,a) = \mathbf{w}^\top \mathbf{Q}(s,a)$, where $Q(s,a)_{i}$ is the Q-value for the $i^{\text{th}}$ objective. $\epsilon$-greedy policy is used on $SQ(s,a)$ during action selection, and regular Q-learning updates are applied on each reward dimension.

\textbf{Mixture Policy} (Mixture) \cite{vamplew2009constructing}: this baseline works by combining multiple Pareto optimal base policies into a single policy. Q-learning is used to optimize for each reward dimension separately (which is approximately Pareto optimal), and the close-to-optimal policy for each dimension is used for $I$ steps before switching to the next. 

\textbf{Welfare Q-learning} (WelfareQ) \cite{fan23}: this baseline extends Q-learning in tabular setting to approximately solve the nonlinear objective function by considering past accumulated rewards to perform non-stationary action selection.

\textbf{Model-Based Mixture Policy} (Mixture-M): Instead of using Q-learning to find an approximately Pareto optimal policy for each objective, value iteration \cite{sutton1998introduction} is used to calculate the optimal policy, and each dimension uses the corresponding optimal policy for $I$ steps before switching to the next.

\subsection{Nonlinear Functions}
\textbf{Taxi}: We use the following three functions on Taxi for fairness considerations: (1) Nash social welfare: $W_{\text{Nash}}(\mathbf{r}) = \left(\prod_{i} r_i \right)^{1/d}$. (2) Egalitarian welfare: $W_{\text{Egalitarian}}(\mathbf{r}) = \min \{ r_i\}_{i}$. (3) $p$-welfare\footnote{$p$-welfare is equivalent to generalized mean. If $p \rightarrow 0$, $p$-welfare converges to Nash welfare; if $p \rightarrow -\infty$, $p$-welfare converges to egalitarian welfare; and $p$-welfare is the utilitarian welfare when $p=1$.}: $W_p(\mathbf{r}) = (\frac{1}{d}\sum_{i} r_{i}^p)^{1/p}$.

\textbf{Scavenger}: We use these two functions to reflect the conflicting nature of the objectives:
\begin{enumerate}[leftmargin=15pt]
    \item Resource-Damage Threshold Scalarization: $RD_{\text{threshold}}(R, D) = R - \max(0, (D - \text{threshold})^3)$, where $R$ is number of resources collected and $D$ is damage taken from enemies. The threshold parameter represents a budget after which the penalty from the damage starts to apply.
    \item Cobb-Douglas Scalarization: $CD_{\rho}(R, D) = R^\rho \left(1/(D + 1)\right)^{(1 - \rho)}$. This function is inspired by economic theory and balances trade-offs between $R$ and $D$.
\end{enumerate}

\subsection{Experiment Settings and Hyperparameters}
We run all the algorithms using 10 random initializations with a fixed seed each. We set $\mathbf{w} = (1/d)\times \mathbf{1}$ (uniform weight) for LinScal, $I = T/d$ for Mixture and Mixture-M, $\alpha=1$ for RAVI, and learning rate of 0.1 for WelfareQ. Three of our baselines (Mixture, LinScal, and WelfareQ) are online algorithms. Thus, to ensure a fair comparison, we tuned their hyperparameters using grid-search and evaluated their performances after they reached convergence. For model-based approaches (RAVI and Mixture-M), we run the algorithms and evaluate them after completion. We set the convergence threshold to $\Delta = 10^{-7}$ for Mixture-M. Some environment-specific settings are discussed below.

\textbf{Taxi}: To ensure numerical stability of $W_{\text{Nash}}$, we optimize its smoothed log-transform $W_{\text{SPF}}(\mathbf{r}, \lambda) = \sum_{i=1}^{d} \ln(r_i + \lambda)$, but during evaluations we still use $W_{\text{Nash}}$. We set $\lambda = 10^{-8}$, $T=100$. $\gamma=1$, size of the grid world to be $15 \times 15$ with $d \in \{2,3,4,5\}$ reward dimensions.

\textbf{Scavenger}: We set threshold for $RD_{\text{threshold}}$ to 2 and $\rho=0.4$ for $CD_{\rho}$. The size of the grid world is $15\times15$ with six resources scattered randomly and 1/3 of the cells randomly populated with enemies. We use $T=20$ and $\gamma=1$.

\begin{table*}[h]
    \centering
    \caption{Comparison with baselines in terms of ESR.}
    \begin{tabular}{@{}lllccccc@{}}
        \toprule
        Environment & Dimension & Function & RAVI & Mixture & LinScal& WelfareQ & Mixture-M \\
        \midrule
         Taxi & $d=2$ & $W_{\text{Nash}}$ & \textbf{7.555$\pm$0.502} & 5.136$\pm$0.547 & 0.000$\pm$0.000 & 5.343$\pm$0.964 & 6.406$\pm$0.628 \\
         & & $W_{\text{Egalitarian}}$ & 3.000$\pm$0.000 & 3.483$\pm$0.894 & 0.000$\pm$0.000 & 0.387$\pm$0.475 & \textbf{4.065$\pm$0.773}\\
         & & $W_{p=-10}$ & \textbf{5.279$\pm$0.000} & 3.623$\pm$0.522 & 0.000$\pm$0.000 & 2.441$\pm$1.518 & 4.356$\pm$0.829\\
         & & $W_{p=0.001}$ & \textbf{7.404$\pm$0.448} & 5.363$\pm$0.844 & 0.000$\pm$0.000 & 4.977$\pm$1.514 & 6.406$\pm$0.628 \\
         & & $W_{p=0.9}$ & \textbf{9.628$\pm$0.349} & 5.947$\pm$0.488 & 7.833$\pm$0.908 & 7.999$\pm$0.822  & 7.052$\pm$0.412\\
         \addlinespace
         & $d=3$ & $W_{\text{Nash}}$ & \textbf{4.996$\pm$0.297} & 3.250$\pm$0.584 & 0.000$\pm$0.000 & 2.798$\pm$1.462 & 3.461$\pm$0.459\\
         & & $W_{\text{Egalitarian}}$ & 2.000$\pm$0.000 & \textbf{2.030$\pm$0.981} & 0.000$\pm$0.000 & 0.094$\pm$0.281 & 1.660$\pm$0.663\\
         & & $W_{p=-10}$ & \textbf{3.115$\pm$0.000} & 2.129$\pm$0.878 & 0.000$\pm$0.000 & 2.558$\pm$0.812 & 1.849$\pm$0.733\\
         & & $W_{p=0.001}$ & 3.307$\pm$0.000 & 3.118$\pm$0.536 & 0.000$\pm$0.000 & 3.306$\pm$1.181 & \textbf{3.462$\pm$0.458}\\
         & & $W_{p=0.9}$ & \textbf{6.250$\pm$0.322} & 3.883$\pm$0.329 & 5.191$\pm$0.464 & 4.834$\pm$0.797 & 4.081$\pm$0.266 \\
         \addlinespace
         & $d=4$ & $W_{\text{Nash}}$ & \textbf{2.191$\pm$0.147} & 0.579$\pm$0.713 & 0.000$\pm$0.000& 1.601$\pm$0.189& 1.122$\pm$0.781\\
         & & $W_{\text{Egalitarian}}$ & \textbf{1.700$\pm$0.483} & 0.545$\pm$0.445 & 0.000$\pm$0.000& 0.000$\pm$0.000 & 0.634$\pm$0.437 \\
         & & $W_{p=-10}$ & \textbf{1.029$\pm$0.000} & 0.490$\pm$0.490 & 0.000$\pm$0.000& 0.689$\pm$0.451& 0.688$\pm$0.475 \\
         & & $W_{p=0.001}$ & \textbf{2.145$\pm$0.129}  & 0.961$\pm$0.628 & 0.000$\pm$0.000& 1.440$\pm$0.511 & 1.126$\pm$0.774 \\
         & & $W_{p=0.9}$ & \textbf{3.369$\pm$0.173} & 1.230$\pm$0.264 & 2.546$\pm$0.174 & 2.521$\pm$0.211 & 1.689$\pm$0.320 \\
         \addlinespace
         & $d=5$ & $W_{\text{Nash}}$ & \textbf{2.308$\pm$0.185} & 0.000$\pm$0.000 & 0.000$\pm$0.000& 1.181$\pm$0.613 & 0.000$\pm$0.000 \\
         & & $W_{\text{Egalitarian}}$ & \textbf{1.700$\pm$0.483} & 0.000$\pm$0.000 & 0.000$\pm$0.000& 0.000$\pm$0.000 & 0.000$\pm$0.000\\
         & & $W_{p=-10}$ & \textbf{1.023$\pm$0.000} & 0.000$\pm$0.000 & 0.000$\pm$0.000 & 0.407$\pm$0.501 & 0.000$\pm$0.000\\
         & & $W_{p=0.001}$ & \textbf{2.000$\pm$0.102} & 0.018$\pm$0.020 & 0.000$\pm$0.000& 1.309$\pm$0.490 & 0.025$\pm$0.025 \\
         & & $W_{p=0.9}$ & \textbf{3.289$\pm$0.147} & 1.220$\pm$0.141 & 2.844$\pm$0.250 & 2.879$\pm$0.263 & 1.761$\pm$0.148 \\
         \addlinespace
          Scavenger & $d=2$ & $CD_{\rho=0.4}$ & \textbf{1.336$\pm$0.240} & 0.655$\pm$0.558 & 0.874$\pm$0.605 & 0.713$\pm$0.645 & 0.729$\pm$0.296\\
          & & $RD_{\text{threshold} = 2}$ & \textbf{3.400$\pm$0.843} & 0.900$\pm$0.943 & 1.200$\pm$0.872 & 1.444$\pm$1.423 & 0.845$\pm$2.938 \\
         \bottomrule
    \end{tabular}
    \label{result:tab1}
\end{table*}
\begin{figure*}[h]
    \begin{subfigure}{0.25\linewidth}
        \centering
        \includegraphics[width=\linewidth]{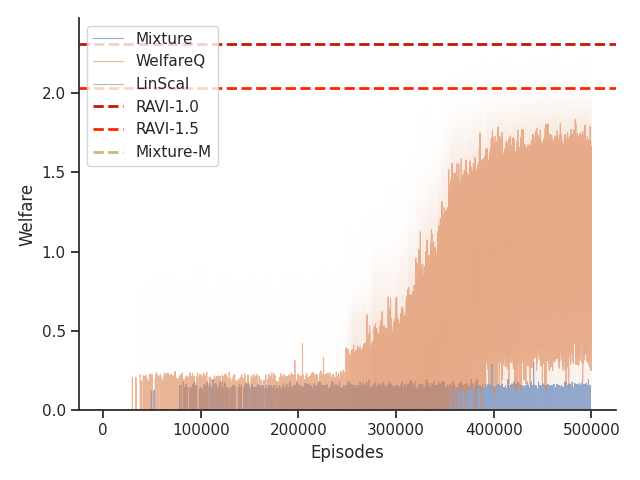}
        \caption*{Taxi, $W_{\text{Nash}}$, $d=5$}
    \end{subfigure}%
    \begin{subfigure}{0.25\linewidth}
        \centering
        \includegraphics[width=\linewidth]{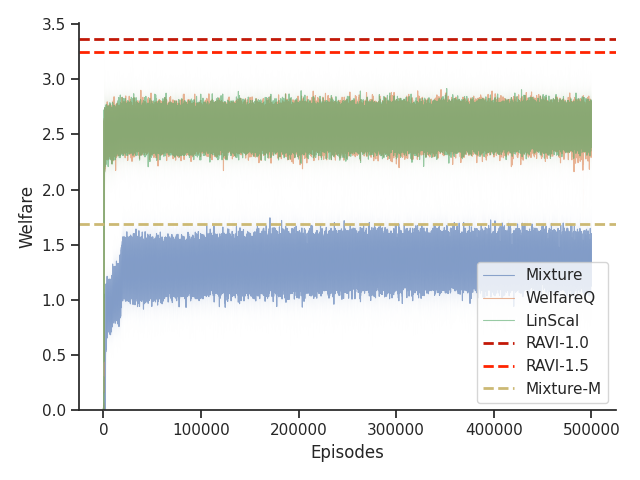}
        \caption*{Taxi, $W_{p=0.9}$, $d=4$}
    \end{subfigure}%
    \begin{subfigure}{0.25\linewidth}
        \centering
        \includegraphics[width=\linewidth]{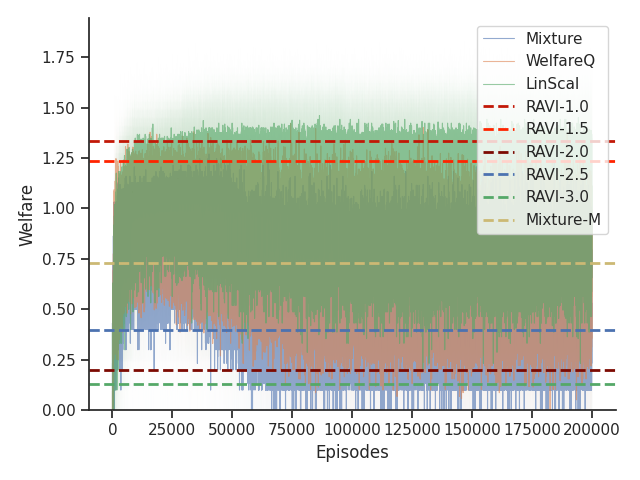}
        \caption*{Scavenger, $CD_{\rho=0.4}$}
    \end{subfigure}%
    \begin{subfigure}{0.25\linewidth}
        \centering
        \includegraphics[width=\linewidth]{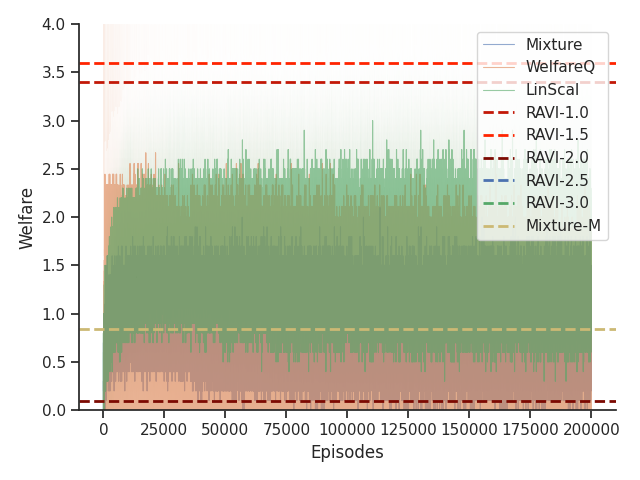}
        \caption*{Scavenger, $RD_{\text{threshold}}$}
    \end{subfigure} \\
    \begin{subfigure}{0.25\linewidth}
        \centering
        \includegraphics[width=\linewidth]{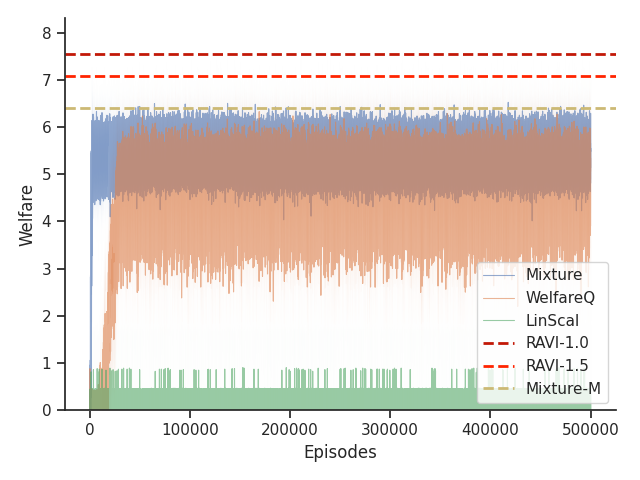}
        \caption*{Taxi, $W_{\text{Nash}}$, $d=2$}
    \end{subfigure}%
    \begin{subfigure}{0.25\linewidth}
        \centering
        \includegraphics[width=\linewidth]{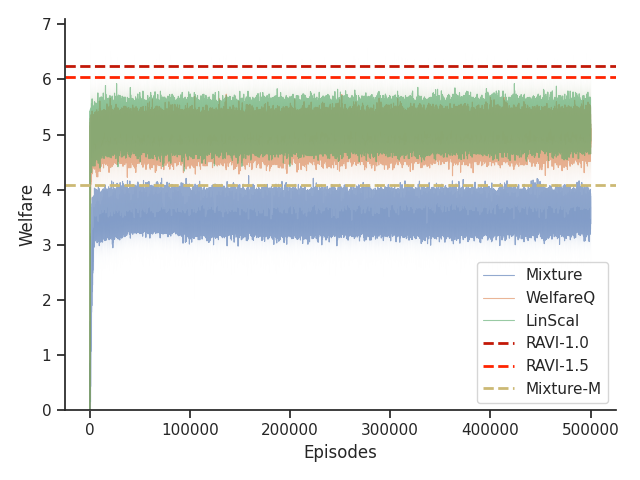}
        \caption*{Taxi, $W_{p=0.9}$, $d=3$}
    \end{subfigure}%
    \begin{subfigure}{0.25\linewidth}
        \centering
        \includegraphics[width=\linewidth]{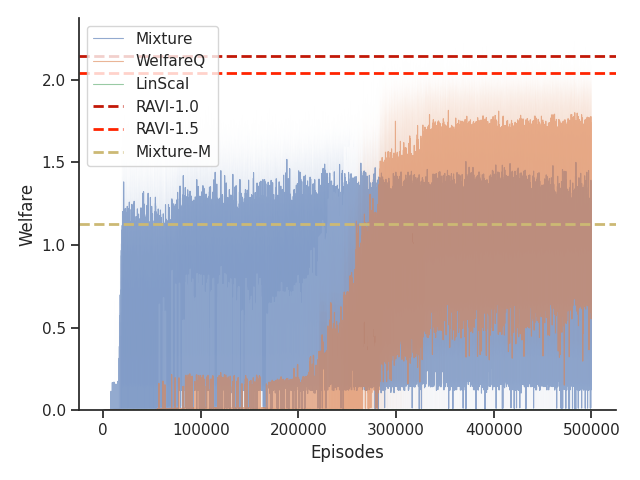}
        \caption*{Taxi, $W_{p=0.001}$, $d=4$}
    \end{subfigure}%
    \begin{subfigure}{0.25\linewidth}
        \centering
        \includegraphics[width=\linewidth]{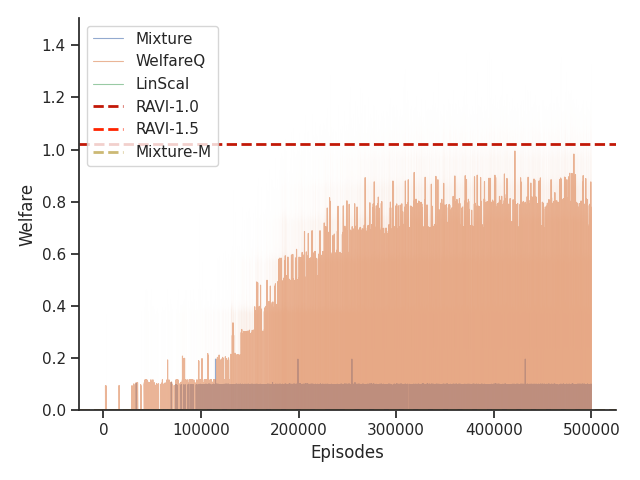}
        \caption*{Taxi, $W_{p=-10}$, $d=5$}
    \end{subfigure} \\
    \caption{Comparisons with baselines, with learning curves included.}
    \label{result:learning curves}
\end{figure*}

\subsection{Results}
As shown in \Cref{result:tab1}, we found that RAVI is able to generally outperform all baselines across all settings in both Taxi and Scavenger environments in terms of optimizing our nonlinear functions of interest.

On the Taxi environment, we observe that LinScal is unable to achieve any performance except for $W_{p=0.9}$. This is an expected behavior due to the use of a linearly scalarized policy and the fact that $p$-welfare converges to utilitarian welfare when $p \rightarrow 1$. Among the baseline algorithms, we observe that WelfareQ performs the best on $d=5$, whereas Mixture and Mixture-M fail due to the use of a fixed interval length for each optimal policy. Furthermore, the advantage of RAVI becomes more obvious as $d$ increases.

On the Scavenger environment, due to $d=2$ and having a dense reward signal, all algorithms are able to achieve reasonable values for the welfare functions. We also observe that RAVI significantly outperforms all baselines.

Given that Mixture, LinScal, and WelfareQ are online algorithms, for the sake of completeness, we provide the visualizations of the learning curves of the baseline algorithms compared with RAVI with different $\alpha$ values in \Cref{result:learning curves}, where the model-base approaches are shown as horizontal dotted lines. In general, we found that the online algorithms tend to converge early, and we observe a degradation in performance as $\alpha$ increases. The full set of results can be found at \Cref{appendix:vis}.

\subsection{Ablation Study on Discretization Factor}
To further evaluate the effect of $\alpha$ on RAVI, we also run two sets of experiments:
\begin{enumerate}[leftmargin=15pt]
    \item we test RAVI performance on settings with $\gamma < 1$ and adopt $\alpha$ values that are substantially greater than those necessary for worst-case theoretical guarantees from the previous analysis. This set of results can be found at \Cref{appendix:gamma leq 1}. Note that the empirical performance of RAVI is substantially better than is guaranteed by the previous theoretical analysis.
    \item With $\gamma = 1$, we evaluate RAVI using larger alpha values and investigate how much performance degradation occurs. This set of experiments can be found in \Cref{appendix:gamma eq 1}.
\end{enumerate}


\section{Removing Knowledge of the Model}
\label{section:remove}
We have shown that \textsc{RAVI} can efficiently find an approximately optimal policy given access to a model of the environment. In this section we observe that the model can be jointly learned by extending the classical E3 algorithm \cite{kearns02} to the nonlinear multi-objective case by lifting the exploration algorithm to the multi-objective setting and using \textsc{RAVI} for the exploitation subroutine. We call the resulting combined algorithm \textsc{Reward-Aware-Explore or Exploit} (or \textsc{RAEE} for short). We briefly explain the high level ideas and state the main result here and defer further discussion to the appendix due to space constraints. 



The algorithm consists of two stages, exploration and exploitation. The algorithm alternates between two stages: exploration and exploitation. Each stage is outlined here at a high level, with more detailed steps provided in the appendix.
\begin{itemize}[leftmargin=10pt]
    \item \textbf{Explore.} At a given state, choose the least experienced action and record the reward and transition. Continue in this fashion until reaching a \textit{known} state, where we say a state is known if it has been visited sufficiently many times for us to have precise local statistics about the reward and transition functions in that state.
    \item \textbf{Exploit.} Run \textsc{RAVI} from the current known state $s$ in the induced model comprising the known states and with a single absorbing state representing all unknown states. If the welfare obtained by this policy is within the desired error bound of $V^*(s, \mathbf{0}, T)$, then we are done. Otherwise, compute a policy to reach an unknown state as quickly as possible and resume exploring.
\end{itemize}

\begin{theorem}[RAEE]
\label{theorem:RA-E3}
Let $V^*(s, 0, T)$ denote the value function of the policy with the optimal expected welfare in the MOMDP $M$ starting at state $s$, with $\mathbf{0} \in \reals^d$ accumulated reward and $T$ timesteps remaining. Then for a uniformly continuous welfare function $W$, there exists an algorithm $A$, taking inputs $\epsilon$, $\beta$, $\mathcal{S}$, $\mathcal{A}$, and $V^*(s, \mathbf{0}, T)$, such that the total number of actions and computation time taken by $A$ is polynomial in $1/\epsilon$, $1/\beta$, $|\mathcal{S}|$, $|\mathcal{A}|$, the horizon time $T = 1/(1 - \gamma)$ and exponential in the number of objectives $d$, and with probability at least $1 - \beta$, $A$ will halt in a state $s$, and output a policy $\hat{\pi}$, such that $V^{\hat{\pi}}_M (s, 0, T) \ge V^*(s, 0, T) -\epsilon$.
\end{theorem}

We provide additional details comparing our analysis with that of \cite{kearns02} in the appendix.

As we do not regard the learning of the Multi-Objective Markov Decision Process (MOMDP) as our primary contribution, we choose to focus the empirical evaluation on the nonlinear optimization subroutine, which is the most crucial modification from the learning problem with a single objective. The environments we used for testing the optimality of RAVI are very interesting for optimizing a nonlinear function of the rewards, but are deterministic in terms of transitions and rewards, making the learning of these environments less interesting empirically.

\section{Conclusion and Future Work}
Nonlinear preferences in reinforcement learning are important, as they can encode \textit{fairness} as the nonlinear balancing of priorities across multiple objectives or \textit{risk attitudes} with respect to even a single objective. Stationary policies are not necessarily optimal for such objectives. We derived an extended form of Bellman optimality to characterize the structure of optimal policies conditional on accumulated reward. We introduced the \textsc{RAVI} and \textsc{RAEE} algorithms to efficiently compute an approximately optimal policy.

Our work is certainly not the first to study MORL including with nonlinear preferences. However, to the best of our knowledge, our work is among the first to provide worst-case approximation guarantees for optimizing ESR in MORL with nonlinear scalarization functions.


While our experiments demonstrate the utility of \textsc{RAVI} in specific settings, there are many possible areas of further empirical evaluation including stochastic environments and model learning alongside the use of \textsc{RAVI} as an exploitation subprocedure as described in theory in Section~\ref{section:remove} with \textsc{RAEE}. Further details on the limitations of our experiments are discussed in \Cref{appendix:exp discussion}.

Our results introduce several natural directions for future work. On the technical side, one could try to handle the case of stochastic reward functions or a large number of objectives. Another direction would involve incorporating function approximation with deep neural networks into the algorithms to enable scaling to even larger state spaces and generalizing between experiences. Our theory suggests that it may be possible to greatly enrich the space of possible policies that can be efficiently achieved in these settings by conditioning function approximators on accumulated reward rather than necessarily considering sequence models over arbitrary past trajectories; we see this as the most exciting next step for applications.

\bibliographystyle{ACM-Reference-Format} 
\bibliography{refs}


\clearpage
\newpage
\appendix
\section*{Appendix}
\section{Proofs of Technical Lemmas}
In this section, we provide proofs of the technical lemmas in Section~\ref{section:model} and Section~\ref{section:algorithm}. 

\subsection{Proof of Lemma \ref{lemma: sufficiency}}
\textbf{Lemma \ref{lemma: sufficiency}}. Let $\mathcal{V}(s, \mathbf{R}(\tau), 0) = W(\mathbf{R}(\tau))$ for all states $s$ and trajectories $\tau$. For every state $s$, history $\tau$, and $t > 0$ time steps remaining, let
    $$
    \mathcal{V}(s, \mathbf{R}(\tau), t) = \max_a \mathbb{E}_{s'} \left[ \mathcal{V} (s', \mathbf{R}(\tau) + \gamma^{T-t}\mathbf{R}(s,a), t-1) \right].
    $$
    Then $V^*(s, \tau, t) = \mathcal{V}(s, \mathbf{R}(\tau), t)$.
    
\begin{proof}
    We proceed by induction on $t$. In the base case of $t=0$, we have simply that $\mathcal{V}(s, \mathbf{R}(\tau), 0) = W(\mathbf{R}(\tau))$ by the definition of $\mathcal{V}$. But when $t=0$, the trajectory has ended, so any $\tau'$ in Definition \ref{definition: Vstar} will be the empty trajectory $\tau_{0;0}$. So 
    $$
    V^*(s, \tau, 0) = W(\mathbf{R}(\tau) + \gamma^{T-t} \mathbf{R}(\tau_{0:0})) = W(\mathbf{R}(\tau)).
    $$

    Suppose the equality holds for all states $s$ and histories $\tau$ for up to $t-1$ time steps remaining, i.e.
    $$
    V^*(s, \tau, t - 1) = \max_a \mathbb{E}_{s'} \left[ \mathcal{V} (s', \mathbf{R}(\tau) + \gamma^{T-t}\mathbf{R}(s,a), t-2) \right].
    $$
    Then with $t$ steps remaining:
    $$
    V^*(s, \tau, t) = \max_{\pi} \sum_{\tau'} Pr^{\pi}(\tau') W(\bm{R}(\tau) + \gamma^{T-t} \bm{R}(\tau'))
    $$
    where the sum is taken over all length-$t$ trajectories $\tau'$ beginning at $s$, and $Pr^{\pi}(\tau')$ is the probability that $\tau'$ is traversed under policy $\pi$. Note that we can further decompose $\tau'$ into $\tau' = (s,a) \oplus \tau''$ where $(s,a)$ denotes the state-action pair corresponding to the current timestep, $\oplus$ concatenates, and $\tau''$ denotes a trajectory of length $t-1$ beginning at some state $s'$ transitioned from state $s$ via taking action $a$. Since the transition to $\tau''$ is independent of earlier states/actions once the current state $s$ and action $a$ are fixed, we can simplify the optimization by shifting the summation from trajectories conditioned on $\pi$ to actions and successor states and thus rewrite $V^*$ as
    \begin{align*}
        V^*(s, \tau, t) = \max_{\pi} \sum_{a} &\pi(a \mid s, \tau, t) 
        \\ \mathbb{E}_{s'} \Bigg[ \sum_{\tau''}Pr^\pi(\tau'')
        & W\big(\bm{R}(\tau) + \gamma^{T-t} \bm{R}(s, a) + \gamma^{T-t+1} \bm{R}(\tau'')\big) \Bigg]
    \end{align*}
    where the latter sum is taken over all length $t-1$ trajectories $\tau''$ beginning at $s'$ and the expectation is taken over the environmental transitions to $s'$ from $s$ given $a$.

    Note that for all actions $a$, $\pi(a \mid s, \tau, t) \in [0,1]$ can be chosen independently of any terms that appears in $Pr^\pi(\tau'')$ for any $\tau''$ by the decomposition of $\tau'$ above. This implies that at the current timestep, the maximizer (or the optimal policy) should choose action $a$ that maximizes the quantity
    $$
    \mathbb{E}_{s'} \bigg[ \sum_{\tau''}Pr^\pi(\tau'') W(\bm{R}(\tau) + \gamma^{T-t} \bm{R}(s, a) + \gamma^{T-t+1} \bm{R}(\tau'')) \bigg]
    $$
    with probability 1. Therefore, the expression for $V^*(s, \tau, t)$ can be rewritten as
    \begin{align*}
        V^*(s, \tau, t) &=\max_{a} \mathbb{E}_{s'} \bigg[ \max_\pi \sum_{\tau''}Pr^\pi(\tau'') W(\bm{R}(\tau) \\
         &\qquad \qquad + \gamma^{T-t} \bm{R}(s, a) + \gamma^{T-(t-1)} \bm{R}(\tau'')) \bigg] \\
        &=\max_{a} \mathbb{E}_{s'} \left[ V^*(s', \tau \oplus (s, a), t-1) \right] \\
        &= \max_{a} \mathbb{E}_{s'} \left[ \mathcal{V}(s', \bm{R}(\tau) + \gamma^{T-t} \bm{R}(s, a), t-1) \right] \\
        &= \mathcal{V}(s, \bm{R}(\tau), t).
    \end{align*}
    where the second equality comes from the definition of $V^*$, the third equality comes from the inductive hypothesis, and the fourth equality comes from the definition of $\mathcal{V}$.
\end{proof}

\subsection{Proof of Lemma \ref{lemma:lipschitz}}

\textbf{Lemma~\ref{lemma:lipschitz}}.
Let the welfare function $W: \mathbb{R}^d \to \mathbb{R}$ be uniformly continuous. Fix $s \in \mathcal{S}$ and $t \in \{0,1,\dots, T\}$, then for all $\epsilon > 0$, there exists $\delta_\epsilon > 0$ such that
$$
\bigg|V^*(s, \mathbf{R_1}, t) - V^*(s, \mathbf{R_2}, t) \bigg| < \epsilon \quad \text{if} \quad \|\mathbf{R_1} - \mathbf{R_2}\| < \delta_\epsilon.
$$

\begin{proof}
    Without loss of generality, assume $V^*(s, \mathbf{R_1}, t) \ge V^*(s, \mathbf{R_2}, t)$. Let $\pi_1$ be the induced optimal policy by $V^*(s, \mathbf{R_1}, t)$, i.e.
    $$
    V^*(s, \mathbf{R_1}, t) = V^{\pi_1}(s, \mathbf{R_1}, t).
    $$
    Then by the optimality of $V^*$,
    \begin{align*}
        V^{\pi_1}(s, \mathbf{R_2}, t) &\le V^*(s, \mathbf{R_2}, t) \\
        &= \max_\pi \sum_\tau Pr^\pi[\tau] W(\mathbf{R_2} + \mathbf{R}(\tau)).
    \end{align*}
    Since $W$ is uniformly continuous, there exists $\delta_\epsilon > 0$ such that $| W(\mathbf{x})-W(\mathbf{y}) | < \epsilon$ if $\|\mathbf{x} - \mathbf{y}\| < \delta_\epsilon$. Then
    \begin{align*}
        &\bigg|V^*(s, \mathbf{R_1}, t) - V^*(s, \mathbf{R_2}, t) \bigg| \\
        \le& \bigg|V^{\pi_1}(s, \mathbf{R_1}, t) - V^{\pi_1}(s, \mathbf{R_2}, t) \bigg| \\
        =& \bigg|\sum_\tau Pr^{\pi_1} [\tau] W(\mathbf{R_1} + \mathbf{R}(\tau)) - \sum_\tau Pr^{\pi_1} [\tau] W(\mathbf{R_2} + \mathbf{R}(\tau)) \bigg| \\
        \le& \sum_\tau Pr^{\pi_1}[\tau] \bigg|W(\mathbf{R_1} + \mathbf{R}(\tau)) - W(\mathbf{R_2} + \mathbf{R}(\tau))\bigg|
    \end{align*}
    where the sum is over all $t$-trajectories $\tau$ that start in state $s$.
    If $\|\mathbf{R_1} - \mathbf{R_2}\| < \delta_\epsilon$ then we have
    $$
    \bigg|W(\mathbf{R_1} + \mathbf{R}(\tau)) - W(\mathbf{R_2} + \mathbf{R}(\tau))\bigg| < \epsilon.
    $$
    Therefore, 
    $$
    \bigg|V^*(s, \mathbf{R_1}, t) - V^*(s, \mathbf{R_2}, t) \bigg| < \epsilon
    $$
    since the sum of probabilities over all $t$-trajectories $\tau$ in $M$ that start in state $s$ induced by $\pi_1$ must add up to $1$.
\end{proof}

\subsection{Proof of Lemma \ref{lemma:error}}

\textbf{Lemma~\ref{lemma:error} (Approximation Error of \textsc{RAVI})}

For uniformly continuous welfare function $W$, for all $\epsilon > 0$, there exists $\alpha_\epsilon$ such that 
    $$
    V(s, \mathbf{R_{acc}}, t) \ge V^*(s, \mathbf{R_{acc}}, t) - t\epsilon
    $$
    \quad $\forall s \in \mathcal{S}, \mathbf{R_{acc}} \in \reals^d, t \in \{0, 1, \dots, T\}$, where $V(s, \mathbf{R_{acc}}, t)$ is computed by Algorithm~\ref{alg:cap} using $\alpha_\epsilon$.
    
\begin{proof}
We prove this additive error bound by induction on the number of timesteps $t$ remaining in the task. 
Clearly $V(s, \mathbf{R_{acc}}, 0) = W(\mathbf{R_{acc}}), \forall s$ and $\mathbf{R_{acc}}$. Suppose that $V(s, \mathbf{R_{acc}}, t) \ge V^*(s, \mathbf{R_{acc}}, t) - t\epsilon$ for all $t = 0, \dots, k - 1$. Then consider with $k$ steps remaining.
\begin{align*} 
V(s, \mathbf{R_{acc}}, k) =& \max_a \sum_{s'}Pr(s'|s, a) V(s', f_\alpha(\mathbf{R_{acc}} + \gamma^{T - k} \mathbf{R}(s,a)), k - 1) \\
\ge& \max_a \sum_{s'}Pr(s'|s, a) \\
& \qquad \bigg( V^* (s', f_\alpha(\mathbf{R_{acc}} +  \gamma^{T - k} \mathbf{R}(s,a)), k - 1 ) - (k - 1)\epsilon \bigg).
\end{align*}

Since $W$ is uniformly continuous, by Lemma \ref{lemma:lipschitz} there exists $\delta_\epsilon$ such that
\begin{align*}
    \bigg| V^* \bigg(s', f_\alpha(\mathbf{R_{acc}} + \gamma^{T - k}& \mathbf{R}(s,a)), k - 1 \bigg) \\
    &- V^* \bigg(s', \mathbf{R_{acc}} +  \gamma^{T - k} \mathbf{R}(s,a), k - 1 \bigg) \bigg| < \epsilon
\end{align*}
if 
$$
\bigg\|f_\alpha(\mathbf{R_{acc}} + \gamma^{T - k} \mathbf{R}(s,a)) - (\mathbf{R_{acc}} +  \gamma^{T - k} \mathbf{R}(s,a)) \bigg\| < \delta_\epsilon.
$$
Hence, it suffices to choose $\alpha_\epsilon = \delta_\epsilon / d$, which implies
\begin{equation*}
    \bigg\|f_\alpha(\mathbf{R_{acc}} + \gamma^{T - k} \mathbf{R}(s,a)) - (\mathbf{R_{acc}} +  \gamma^{T - k} \mathbf{R}(s,a)) \bigg\| < (\delta_\epsilon / d) \cdot d = \delta_\epsilon,
\end{equation*}
and so
\begin{align*}
    V(s, \mathbf{R_{acc}}, k)
    \ge& \max_a \sum_{s'}Pr(s'|s, a) \\
    & \qquad V^* \bigg(s', \mathbf{R_{acc}} +  \gamma^{T - k} \mathbf{R}(s,a), k - 1 \bigg) - \epsilon - (k - 1)\epsilon \\
    =& V^*(s, \mathbf{R}_{acc}, k) - k\epsilon.
\end{align*}
\end{proof}

\subsection{Lemma \ref{lemma:horizon} (Horizon Time)}

\begin{lemma}
\label{lemma:horizon}
    Let $M$ be any MOMDP, and let $\pi$ be any policy in $M$. Assume the welfare function $W: \mathbb{R}^d \to \mathbb{R}$ is uniformly continuous and $\mathbf{R}(s,a) \in [0,1]^d ~ \forall s, a$. Then for all $\epsilon > 0$, there exists $\delta_\epsilon > 0$ such that if
    $$
    T \ge \left(\frac1{1-\gamma}\right)\log\left(\frac{\sqrt{d}}{\delta_\epsilon(1-\gamma)}\right)
    $$
    then for any state $s,$
    $$
    V^\pi(s, \mathbf{0}, T) \le \lim_{T \to \infty} V^\pi(s, \mathbf{0}, T) \le V^\pi(s, \mathbf{0}, T) + \epsilon.
    $$
    
    We call the value of the lower bound on $T$ given above the $\epsilon$-horizon time for the MOMDP $M$.
\end{lemma}

\begin{proof}
The lower bound on $\lim_{T \to \infty} V^\pi(s, \mathbf{0}, T)$ follows from the definitions, since all rewards are nonnegative and the welfare function should be monotonic. For the upper bound, fix any infinite trajectory $\tau$ that starts at $s$ and let $\tau'$ be the $T$-trajectory prefix of the infinite trajectory $\tau$ for some finite $T$. Since $W$ is uniformly continuous, for all $\epsilon > 0$, there exists $\delta_\epsilon > 0$ such that $W(\mathbf{R}(\tau)) \le W(\mathbf{R}(\tau')) + \epsilon$ if $\|\sum_{k=1}^{\infty} \gamma^{k-1} \bm{R}(s_k, a_k) - \sum_{k=1}^{T} \gamma^{k-1} \bm{R}(s_k, a_k) \| < \delta_\epsilon$, where $\|\cdot\|$ denotes the usual Euclidean norm. But
\begin{align*}
&\bigg\|\sum_{k=1}^{\infty} \gamma^{k-1} \bm{R}(s_k, a_k) - \sum_{k=1}^{T} \gamma^{k-1} \bm{R}(s_k, a_k) \bigg\| \\
=& \bigg\|\sum_{k=T+1}^{\infty} \gamma^{k-1} \bm{R}(s_k,a_k)  \bigg
\| \\
=& \gamma^T\bigg\|\sum_{k=1}^{\infty} \gamma^{k-1} \bm{R}(s_{T+k}, a_{T+k})\bigg\| \\
\le& \gamma^T\left(\frac{\sqrt{d}}{1-\gamma}\right)
\end{align*}
Solving 
$$
\gamma^T\left(\frac{\sqrt{d}}{1-\gamma}\right) \le \delta_\epsilon
$$
for $T$ yields the desired bound on $T$. Since the inequality holds for every fixed trajectory, it also holds for the distribution over trajectories induced by any policy $\pi$.
\end{proof}

\section{Necessity of Conditioning on Remaining Timesteps \( t \)}

Our opening example in Figure~\ref{fig:example} mentioned the necessity of conditioning on accumulated reward to model the optimal policy. However, the formulation of the optimal value function in Definition~\ref{definition:recvstar} also conditions on the number of time-steps remaining in the finite-time horizon task. One may naturally see this as undesirable in terms of adding computational overhead. Particularly in the continuing task setting with $\gamma$ very close to 1, it is natural to ask whether this conditioning is necessary.

\begin{lemma}
In MORL with a nonlinear scalarization function \( W \), for any constant \( \alpha > 0 \), there exists an MOMDP such that any policy \( \pi \) that depends only on the current state and accumulated reward (and does not condition on the remaining timesteps \( t \)) achieves expected welfare \( V^{\pi} \leq \alpha V^{\pi^*} \), where \( V^{\pi^*} \) is the expected welfare achieved by an optimal policy \( \pi^* \) that conditions on \( t \).
\end{lemma}

\begin{proof}
We construct an MOMDP where the optimal action at a decision point depends critically on the remaining timesteps \( t \). We show that any policy not conditioning on \( t \) cannot, in general, approximate the optimal expected welfare within any constant factor.

Consider an MDP with states \( s_0 \), \( s_1 \), and \( s_2 \), where the agent chooses between actions \( a \) and \( b \) at \( s_0 \) and before reaching terminal states $s1$ or $s2$ with no further actions or rewards after reaching them. Choosing action \( a \) results in a reward \( \mathbf{R}(s_0, a) = (\epsilon, R) \) and transition to \( s_1 \). Choosing action \( b \) results in a reward \( \mathbf{R}(s_0, b) = (2\epsilon, 0) \) and transition to \( s_2 \). The accumulated reward prior to reaching \( s_0 \) is \( \mathbf{R}_{\text{acc}} = (0, D) \), where \( D > 0 \). The scalarization function is \( W(x_1, x_2) = x_1 + x_2^2 \), and we assume a discount factor \( \gamma \in (0,1) \).

The agent must choose between actions \( a \) and \( b \) at \( s_0 \) to maximize the expected welfare:
\[
V = W\left( \mathbf{R}_{\text{acc}} + \gamma^{T - t} \mathbf{R}(s_0, a \text{ or } b) \right).
\]

For action \( a \):
\begin{align*}
    \mathbf{R}^{(a)}_{\text{total}} &= \mathbf{R}_{\text{acc}} + \gamma^{T - t} \mathbf{R}(s_0, a) = \left( \gamma^{T - t} \epsilon, D + \gamma^{T - t} R \right), \\
    V^{(a)} &= W\left( \mathbf{R}^{(a)}_{\text{total}} \right) = \gamma^{T - t} \epsilon + \left( D + \gamma^{T - t} R \right)^2.
\end{align*}

For action \( b \):
\begin{align*}
    \mathbf{R}^{(b)}_{\text{total}} &= \mathbf{R}_{\text{acc}} + \gamma^{T - t} \mathbf{R}(s_0, b) = \left( 2 \gamma^{T - t} \epsilon, D \right), \\
    V^{(b)} &= W\left( \mathbf{R}^{(b)}_{\text{total}} \right) = 2 \gamma^{T - t} \epsilon + D^2.
\end{align*}

The difference in expected welfare is given by:
\[
\Delta V = V^{(a)} - V^{(b)} = \gamma^{T - t} \left( -\epsilon + 2 D R + \gamma^{T - t} R^2 \right).
\]

When $T - t \to 0$ (i.e., \( t \) is close to \( T \), near the end of the horizon):
\[
\gamma^{T - t} \to 1, \quad \text{and} \quad \Delta V \to  -\epsilon + 2 D R + R^2.
\]
Since \( R \) is large, \( 2 D R + R^2 \) dominates, and \( \Delta V > 0 \), making action \( a \) optimal.

When \( T - t \) is large (i.e., \( t \) is small, far from \( T \)):
\[
\gamma^{T - t} \ll 1, \quad \text{and} \quad \Delta V \to \gamma^{T - t} \left( -\epsilon + 2 D R \right).
\]

If we choose \( D = \frac{\epsilon - \delta}{2 R} \) for some small \( \delta > 0 \), then:

  \[
  -\epsilon + 2 D R = -\epsilon + 2 \left( \frac{\epsilon - \delta}{2 R} \right) R = -\delta < 0.
  \]

Thus, \( \Delta V < 0 \), making action \( b \) optimal.

Any policy \( \pi \) that does not condition on \( t \) must choose either action \( a \) or \( b \) at \( s_0 \) regardless of \( t \). If \( \pi \) always chooses \( a \), it will be suboptimal for large \( T - t \), when action \( b \) is optimal. If \( \pi \) always chooses \( b \), it will be suboptimal for small \( T - t \), when action \( a \) is optimal.

Let \( \pi^* \) be the optimal policy that conditions on \( t \). When action \( a \) is optimal, we have
$$
\frac{V^{\pi}}{V^{\pi^*}} = \frac{2 \gamma^{T - t} \epsilon + D^2}{\gamma^{T - t} \epsilon + \left( D + \gamma^{T - t} R \right)^2}.
$$

As \( R \to \infty \), \( V^{\pi^*} \to \left( D + \gamma^{T - t} R \right)^2 \), which grows quadratically with \( R \), while \( V^{\pi} \) remains bounded. Therefore, the ratio approaches zero:

  \[
  \lim_{R \to \infty} \frac{V^{\pi}}{V^{\pi^*}} = 0.
  \]

When action \( b \) is optimal, the analysis is analogous, leading to a similar conclusion.

Thus, for any constant \( \alpha > 0 \), we can choose parameters \( \epsilon, R, D, \gamma, T \) such that the expected welfare ratio \( \frac{V^{\pi}}{V^{\pi^*}} \leq \alpha \).
\end{proof}

\section{Ablation Studies and Additional Experiments}

\subsection{Ablation Studies with $\gamma < 1$}\label{appendix:gamma leq 1}

\begin{figure*}
    \begin{minipage}{\textwidth}
        \begin{subfigure}{0.5\textwidth}
            \centering
            \includegraphics[width=\textwidth]{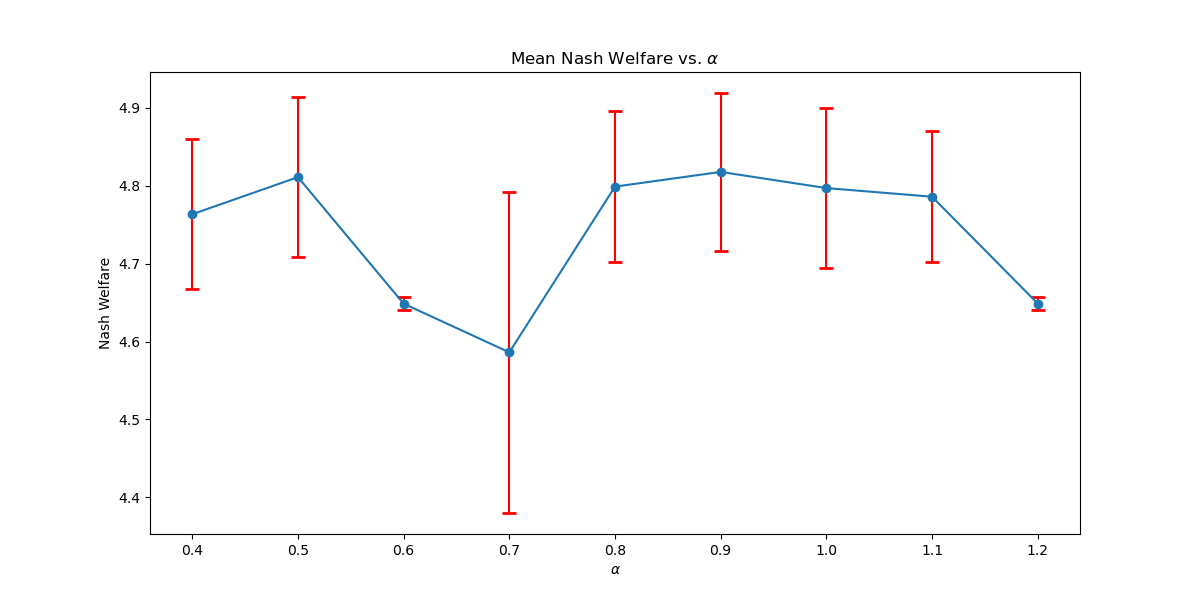}
            \caption{Nash Welfare in Taxi}
            \label{result:ablation:a}
        \end{subfigure}%
        \begin{subfigure}{0.5\textwidth}
            \centering
            \includegraphics[width=\textwidth]{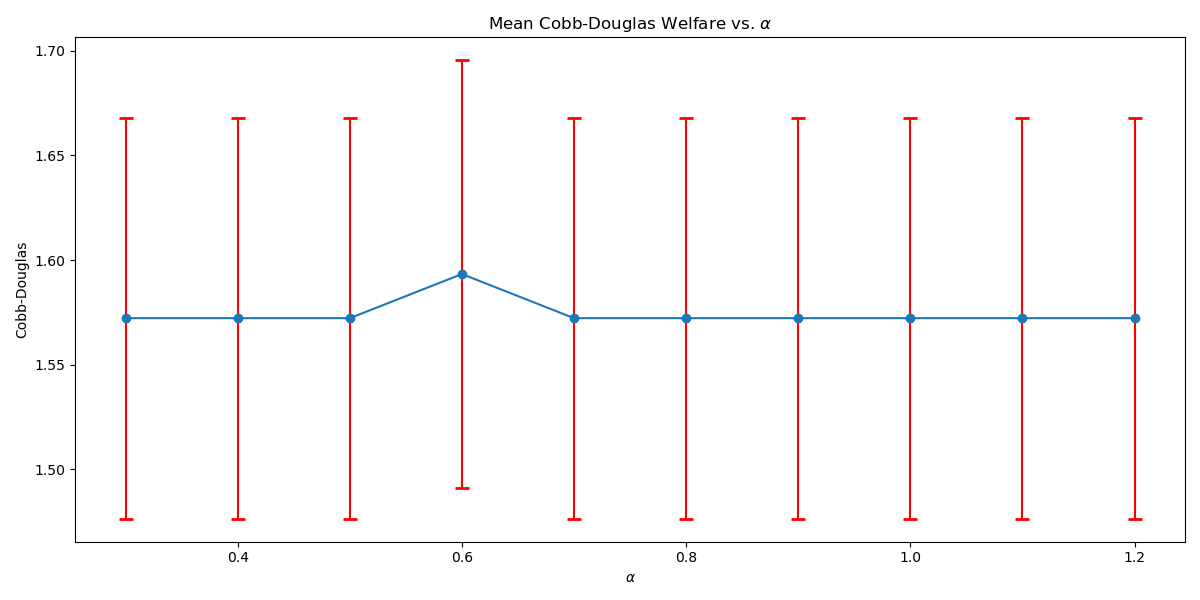}
            \caption{Cobb-Douglas in Scavenger Hunt}
            \label{result:ablation:b}
        \end{subfigure}%
        \caption{Ablation study on discretization factor $\alpha$.}
        \label{appendix:ablation}
    \end{minipage}
\end{figure*}

We conduct further studies on the empirical effects of the discretization factor $\alpha$. For $W_{\text{SPF}}$, we set $\lambda = 1$ (the smoothing factor). We gradually increase the disctretization interval of accumulated reward $\alpha$ gradually from $0.4$ to $1.2$ for a reward discount $\gamma = 0.999$ for Taxi and $\gamma = 0.99$ for Scavenger Hunt with the same settings as the experiments. As shown in \Cref{result:ablation:a} and \Cref{result:ablation:b}, we observe that the solution quality is often very high (better than baselines considered in the main body) even for discretization values that are substantially greater than those needed for theoretical guarantees (with $\epsilon = 1$, we need $\alpha = 2.5\times10^{-3}$). Nonetheless, the results exhibit non-monotonic behavior with respect to $\alpha$, which suggests that factors such as the alignment of discretization granularity with the welfare function's smoothness play a role, especially in this large $\alpha$ regime where empirical performance exceeds the worst-case theoretical guarantees.


This set of ablation studies demonstrates that the \textsc{RAVI} algorithm is capable of finding high-quality policies for optimizing expected scalarized return despite using a larger discretization parameter than that needed for our theoretical analysis. The practical takeaway is that, while smaller $\alpha$ values provide finer discretization and better theoretical guarantees, they incur increased computation costs. Conversely, larger $\alpha$ values can coarsen the discretization but still result in competitive performance, as demonstrated empirically.


\subsection{Ablation Studies with $\gamma = 1$}\label{appendix:gamma eq 1}
We also conduct further ablation studies on environments with $\gamma = 1$ using coarser alpha values. We select $\alpha \in \{1, 1.5, 2, 2.5, 3\}$ and run 10 random initialization on each environment for each value. Note that the smallest $\alpha$ possible in both Taxi and Scavenger is $\alpha = 1$ (which are the results we reported in the main text). Our results can be found at \Cref{result:ablation:coarse alphas taxi} and \Cref{result:ablation:coarse alphas scavenger}. We observe that there is a noticeable performance drop across both environments with a larger $\alpha$. Moreover, due to the nature of having a sparse reward signal in the Taxi environment, we observe no performance when $\alpha \geq 2.0$. On Scavenger, for $RD_{\text{threshold=2}}$, we observe that the agent is under-performing when $\alpha \geq 2.5$. This finding is also to be expected since the resources are scarce (only 6 resources in a $15\times15$ grid), and enemies are plenty (33\% of the grid world).
\begin{table*}[h]
    \caption{Ablation study on discretization factor $\alpha$ with $\gamma = 1$ on Taxi.}
    \centering
    \begin{tabular}{lllccccc}
        \toprule
         Dimension & Alpha & $W_{\text{Nash}}$ & $W_{\text{Egalitarian}}$ & $W_{p=-10}$ & $W_{p=0.001}$ & $W_{p=0.9}$  \\
         \midrule
         $d=2$ & $\alpha=1.0$ &7.555$\pm$0.502 & 3.000$\pm$0.000 & 5.279$\pm$0.000 & 7.404$\pm$0.448 & 9.628$\pm$0.349 \\
         &$\alpha = 1.5$ & 7.079$\pm$0.318 & 1.000$\pm$0.000 & 3.198$\pm$0.000 & 6.481$\pm$0.000 & 8.877$\pm$0.528 \\
         & $\alpha = 2.0$ & 0.000$\pm$0.000 & 0.000$\pm$0.000 & 0.000$\pm$0.000 & 0.000$\pm$0.000 & 0.000$\pm$0.000 \\
         & $\alpha = 2.5$ & 0.000$\pm$0.000 & 0.000$\pm$0.000 & 0.000$\pm$0.000 & 0.000$\pm$0.000 & 0.000$\pm$0.000 \\
         & $\alpha = 3.0$ & 0.000$\pm$0.000 & 0.000$\pm$0.000 & 0.000$\pm$0.000 & 0.000$\pm$0.000 & 0.000$\pm$0.000 \\
         \addlinespace
         $d=3$ & $\alpha=1.0$ & 4.996$\pm$0.297 & 2.000$\pm$0.000 & 3.115$\pm$0.000 & 3.307$\pm$0.000 & 6.250$\pm$0.322 \\
         & $\alpha=1.5$ & 4.902$\pm$0.273 & 1.000$\pm$0.000 & 1.116$\pm$0.000 & 2.080$\pm$0.000 & 6.037$\pm$0.322 \\
         & $\alpha = 2.0$ & 0.000$\pm$0.000 & 0.000$\pm$0.000 & 0.000$\pm$0.000 & 0.000$\pm$0.000 & 0.000$\pm$0.000 \\
         & $\alpha = 2.5$ & 0.000$\pm$0.000 & 0.000$\pm$0.000 & 0.000$\pm$0.000 & 0.000$\pm$0.000 & 0.000$\pm$0.000 \\
         & $\alpha = 3.0$ & 0.000$\pm$0.000 & 0.000$\pm$0.000 & 0.000$\pm$0.000 & 0.000$\pm$0.000 & 0.000$\pm$0.000 \\
         \addlinespace
         $d=4$ & $\alpha=1.0$ & 2.191$\pm$0.147 & 1.700$\pm$0.483 & 1.029$\pm$0.000 & 2.145$\pm$0.129 & 3.369$\pm$0.173 \\
         & $\alpha=1.5$ & 2.044$\pm$0.157 & 1.000$\pm$0.000 & 0.000$\pm$0.000 & 2.044$\pm$0.157 & 3.247$\pm$0.183 \\
         & $\alpha = 2.0$ & 0.000$\pm$0.000 & 0.000$\pm$0.000 & 0.000$\pm$0.000 & 0.000$\pm$0.000 & 0.000$\pm$0.000 \\
         & $\alpha = 2.5$ & 0.000$\pm$0.000 & 0.000$\pm$0.000 & 0.000$\pm$0.000 & 0.000$\pm$0.000 & 0.000$\pm$0.000 \\
         & $\alpha = 3$ & 0.000$\pm$0.000 & 0.000$\pm$0.000 & 0.000$\pm$0.000 & 0.000$\pm$0.000 & 0.000$\pm$0.000 \\
         \addlinespace
         $d=5$& $\alpha=1.0$ & 2.308$\pm$0.185 & 1.700$\pm$0.483 & 1.023$\pm$0.000 & 2.000$\pm$0.102 & 3.289$\pm$0.147 \\
         & $\alpha=1.5$ & 2.034$\pm$0.122 & 1.000$\pm$0.000 & 0.000$\pm$0.000 & 1.149$\pm$0.000 & 3.097$\pm$0.139 \\
         & $\alpha = 2.0$ & 0.000$\pm$0.000 & 0.000$\pm$0.000 & 0.000$\pm$0.000 & 0.000$\pm$0.000 & 0.000$\pm$0.000 \\
         & $\alpha = 2.5$ & 0.000$\pm$0.000 & 0.000$\pm$0.000 & 0.000$\pm$0.000 & 0.000$\pm$0.000 & 0.000$\pm$0.000 \\
         & $\alpha = 3.0$ & 0.000$\pm$0.000 & 0.000$\pm$0.000 & 0.000$\pm$0.000 & 0.000$\pm$0.000 & 0.000$\pm$0.000 \\
         \bottomrule
    \end{tabular}
    \label{result:ablation:coarse alphas taxi}
\end{table*}
\begin{table*}
    \caption{Ablation study on discretization factor $\alpha$ with $\gamma=1$ on Scavenger.}
    \centering
    \begin{tabular}{llcc}
    \toprule
         Dimension & Alpha & $CD_{\rho=0.4}$ & $RD_{\text{threshold} = 2}$  \\
    \midrule
         $d=2$ & $\alpha=1.0$ & 1.336$\pm$0.240 & 3.400$\pm$0.843 \\ 
         & $\alpha=1.5$ & 1.236$\pm$0.279 & 3.600$\pm$0.699 \\
         & $\alpha=2.0$ & 0.197$\pm$0.381 & 0.100$\pm$0.316 \\
         & $\alpha=2.5$ & 0.400$\pm$0.578 & -1.889$\pm$3.480 \\
         & $\alpha=3.0$ & 0.132$\pm$0.295 & -12.500$\pm$22.405 \\
    \bottomrule
    \end{tabular}
    \label{result:ablation:coarse alphas scavenger}
\end{table*}
\begin{table*}
    \caption{Runtime comparisons (in real-time hours).}
    \centering
    \begin{tabular}{lllccccc}
    \toprule
         Environment & Dimension & Function & RAVI & Mixture &  LinScal & WelfareQ & Mixture-M  \\
    \midrule
         Taxi & $d = 2$ & $W_\text{Nash}$ & 0.03 & 0.76 $\pm$ 0.20 & 1.98 $\pm$ 0.28 & 4.82 $\pm$ 1.65 & 0.15 \\
        & & $W_\text{Egalitarian}$ & 0.01 & 0.69 $\pm$ 0.22 & 2.00 $\pm$ 0.31 & 2.61 $\pm$ 0.50 & 0.15 \\
        & & $W_{p=-10}$ & 0.01 & 0.70 $\pm$ 0.24 & 2.11 $\pm$ 0.09 & 4.72 $\pm$ 1.81 & 0.15 \\
        & & $W_{p=0.001}$ & 0.02 & 0.65 $\pm$ 0.29 & 2.10 $\pm$ 0.09 & 5.10 $\pm$ 1.75 & 0.15 \\
        & & $W_{p=0.9}$ & 0.08 & 0.64 $\pm$ 0.21 & 2.11 $\pm$ 0.09 & 4.96 $\pm$ 1.77 & 0.15 \\
        \addlinespace
        & $d = 3$ & $W_\text{Nash}$ & 0.30 & 0.60 $\pm$ 0.29 & 2.14 $\pm$ 0.39 & 4.92 $\pm$ 1.91 & 0.32 \\
        & & $W_\text{Egalitarian}$ & 0.01 & 0.72 $\pm$ 0.21 & 2.09 $\pm$ 0.30 & 2.26 $\pm$ 0.58 & 0.32 \\
        & & $W_{p=-10}$ & 0.02 & 0.64 $\pm$ 0.27 & 2.18 $\pm$ 0.06 & 5.44 $\pm$ 1.91 & 0.32 \\
        & & $W_{p=0.001}$ & 0.02 & 0.75 $\pm$ 0.11 & 2.10 $\pm$ 0.26 & 5.73 $\pm$ 1.80 & 0.32 \\
        & & $W_{p=0.9}$ & 2.10 & 0.74 $\pm$ 0.28 & 2.18 $\pm$ 0.11 & 5.47 $\pm$ 1.79 & 0.32 \\
        \addlinespace
        & $d = 4$ & $W_\text{Nash}$ & 0.65 & 0.82 $\pm$ 0.14 & 2.19 $\pm$ 0.13 & 5.44 $\pm$ 1.62 & 0.53 \\
        & & $W_\text{Egalitarian}$ & 0.06 & 0.71 $\pm$ 0.33 & 2.18 $\pm$ 0.13 & 2.47 $\pm$ 0.64 & 0.53 \\
        & & $W_{p=-10}$ & 0.02 & 0.69 $\pm$ 0.20 & 2.12 $\pm$ 0.31 & 5.66 $\pm$ 1.58 & 0.53 \\
        & & $W_{p=0.001}$ & 0.23 & 0.72 $\pm$ 0.23 & 2.21 $\pm$ 0.11 & 5.11 $\pm$ 1.80 & 0.53 \\
        & & $W_{p=0.9}$ & 36.72 & 0.73 $\pm$ 0.22 & 2.21 $\pm$ 0.07 & 6.15 $\pm$ 1.78 & 0.53 \\
        \addlinespace
        & $d = 5$ & $W_\text{Nash}$ & 1.46 & 0.70 $\pm$ 0.21 & 2.25 $\pm$ 0.10 & 4.93 $\pm$ 1.72 & 0.75 \\
        & & $W_\text{Egalitarian}$ & 0.28 & 0.71 $\pm$ 0.18 & 2.23 $\pm$ 0.05 & 2.37 $\pm$ 0.51 & 0.75 \\
        & & $W_{p=-10}$ & 0.04 & 0.82 $\pm$ 0.08 & 2.21 $\pm$ 0.09 & 5.40 $\pm$ 1.83 & 0.75 \\
        & & $W_{p=0.001}$ & 0.26 & 0.68 $\pm$ 0.17 & 2.05 $\pm$ 0.41 & 5.45 $\pm$ 1.77 & 0.75 \\
        & & $W_{p=0.9}$ & 118.63 & 0.67 $\pm$ 0.27 & 2.16 $\pm$ 0.30 & 4.75 $\pm$ 1.59 & 0.75 \\
        \addlinespace
        Scavenger & $d = 2$ & $CD_{p=0.4}$ & 0.11 & 0.30 $\pm$ 0.05 & 0.25 $\pm$ 0.06 & 0.34 $\pm$ 0.05 & 4.71 \\
        & & $RD_\text{threshold=2}$ & 0.09 & 0.25 $\pm$ 0.06 & 0.30 $\pm$ 0.05 & 0.29 $\pm$ 0.06 & 4.39 \\ 
    \bottomrule
    \end{tabular}
    \label{result:runtime comparison}
\end{table*}

\subsection{Visualizations of RAVI and Baselines}\label{appendix:vis}
\begin{figure*}[h]
    \begin{subfigure}{0.25\linewidth}
        \centering
        \includegraphics[width=\linewidth]{figs/taxi-nash-welfare-15-2.png}
        \caption*{$d=2$}
    \end{subfigure}%
    \begin{subfigure}{0.25\linewidth}
        \centering
        \includegraphics[width=\linewidth]{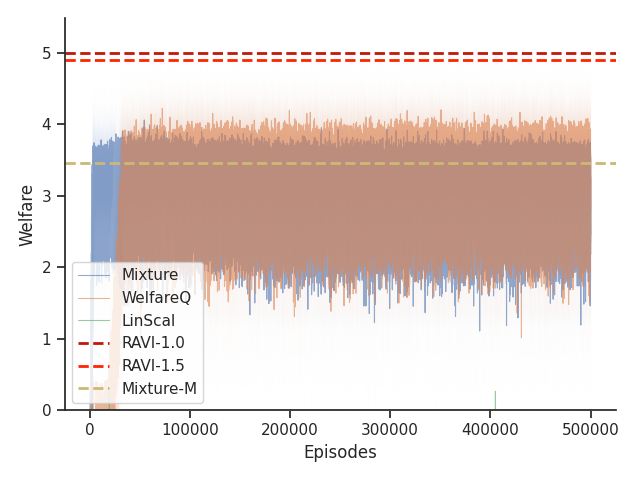}
        \caption*{$d=3$}
    \end{subfigure}%
    \begin{subfigure}{0.25\linewidth}
        \centering
        \includegraphics[width=\linewidth]{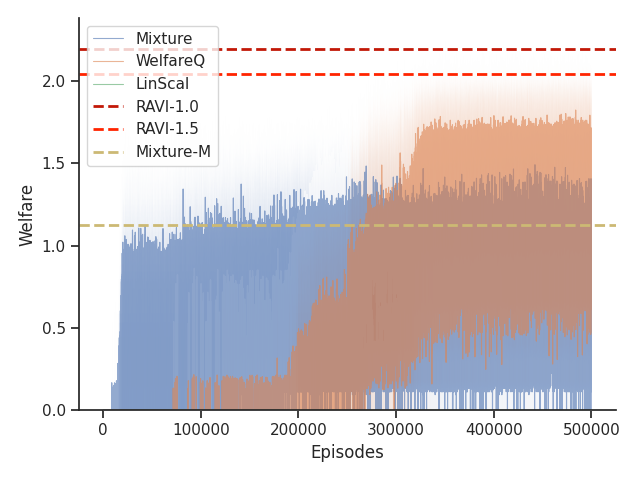}
        \caption*{$d=4$}
    \end{subfigure}%
    \begin{subfigure}{0.25\linewidth}
        \centering
        \includegraphics[width=\linewidth]{figs/taxi-nash-welfare-15-5.png}
        \caption*{$d=5$}
    \end{subfigure}%
    \caption{Taxi, $W_{\text{Nash}}$}
    \label{appendix:learning curve taxi nash}
\end{figure*}

\begin{figure*}[h]
    \begin{subfigure}{0.25\linewidth}
        \centering
        \includegraphics[width=\linewidth]{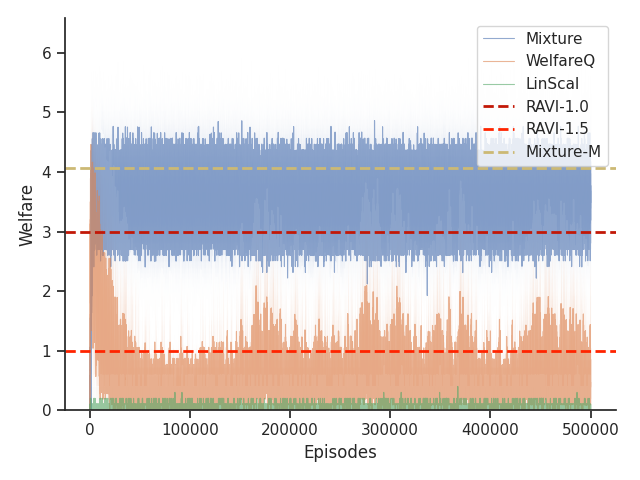}
        \caption*{$d=2$}
    \end{subfigure}%
    \begin{subfigure}{0.25\linewidth}
        \centering
        \includegraphics[width=\linewidth]{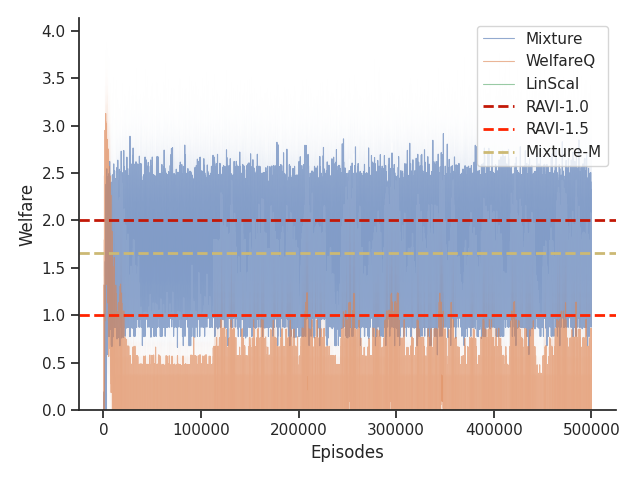}
        \caption*{$d=3$}
    \end{subfigure}%
    \begin{subfigure}{0.25\linewidth}
        \centering
        \includegraphics[width=\linewidth]{figs/taxi-egalitarian-15-4.png}
        \caption*{$d=4$}
    \end{subfigure}%
    \begin{subfigure}{0.25\linewidth}
        \centering
        \includegraphics[width=\linewidth]{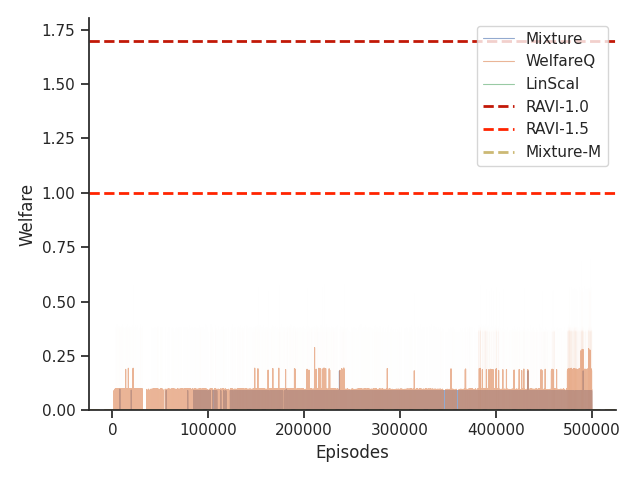}
        \caption*{$d=5$}
    \end{subfigure}%
    \caption{Taxi, $W_{\text{Egalitarian}}$}
    \label{appendix:learning curve taxi egalitarian}
\end{figure*}

\begin{figure*}[h]
    \begin{subfigure}{0.25\linewidth}
        \centering
        \includegraphics[width=\linewidth]{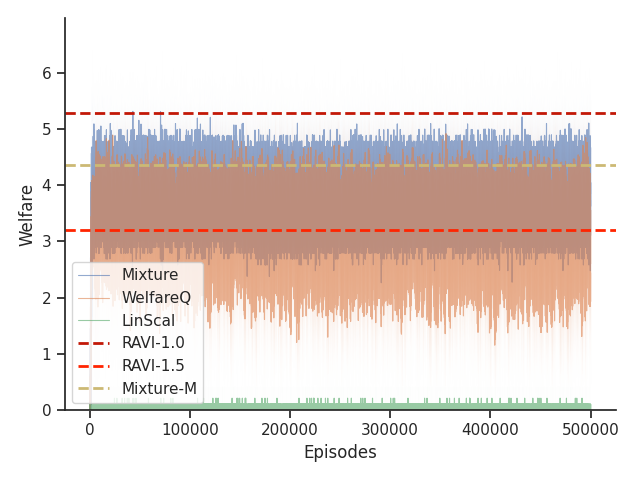}
        \caption*{$d=2$}
    \end{subfigure}%
    \begin{subfigure}{0.25\linewidth}
        \centering
        \includegraphics[width=\linewidth]{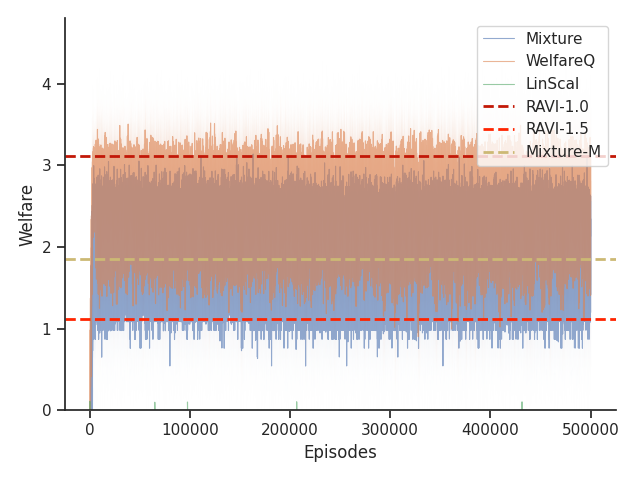}
        \caption*{$d=3$}
    \end{subfigure}%
    \begin{subfigure}{0.25\linewidth}
        \centering
        \includegraphics[width=\linewidth]{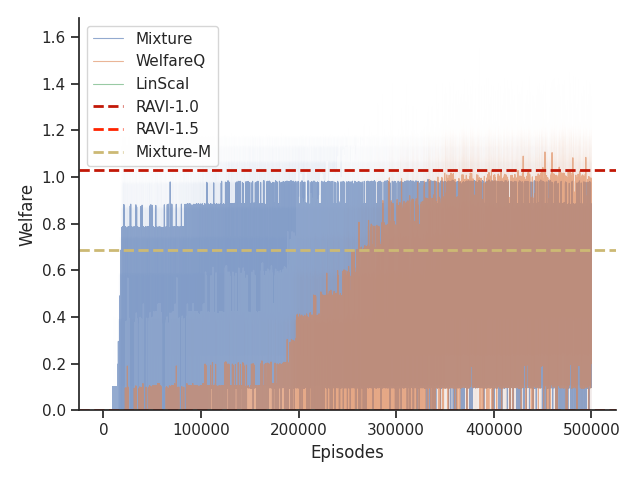}
        \caption*{$d=4$}
    \end{subfigure}%
    \begin{subfigure}{0.25\linewidth}
        \centering
        \includegraphics[width=\linewidth]{figs/taxi-p-welfare--10-15-5.png}
        \caption*{$d=5$}
    \end{subfigure}%
    \caption{Taxi, $W_{p=-10}$}
    \label{appendix:learning curve taxi p=-10}
\end{figure*}

\begin{figure*}[h]
    \begin{subfigure}{0.25\linewidth}
        \centering
        \includegraphics[width=\linewidth]{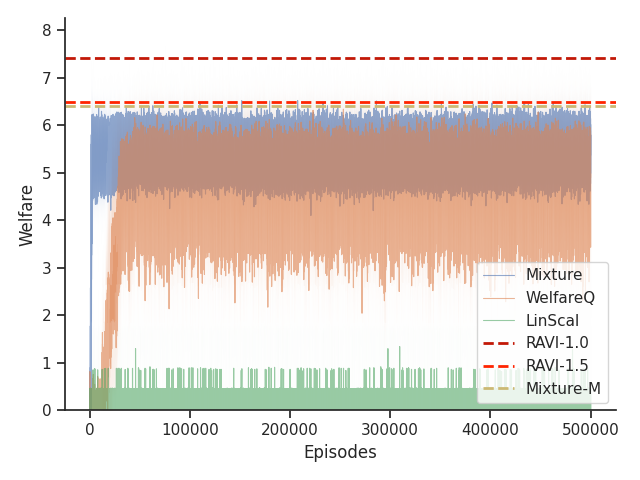}
        \caption*{$d=2$}
    \end{subfigure}%
    \begin{subfigure}{0.25\linewidth}
        \centering
        \includegraphics[width=\linewidth]{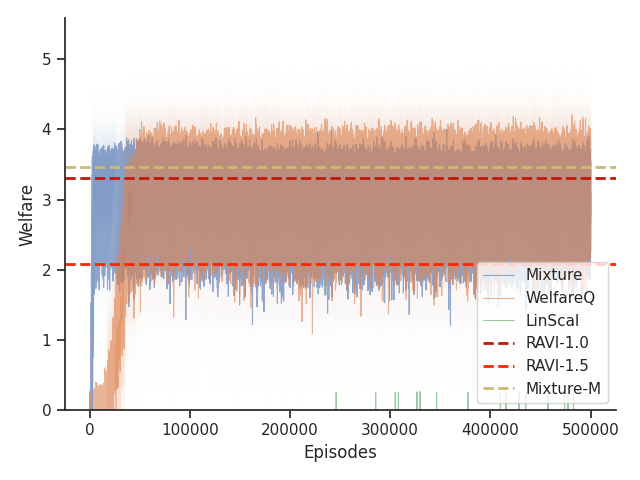}
        \caption*{$d=3$}
    \end{subfigure}%
    \begin{subfigure}{0.25\linewidth}
        \centering
        \includegraphics[width=\linewidth]{figs/taxi-p-welfare-0.001-15-4.png}
        \caption*{$d=4$}
    \end{subfigure}%
    \begin{subfigure}{0.25\linewidth}
        \centering
        \includegraphics[width=\linewidth]{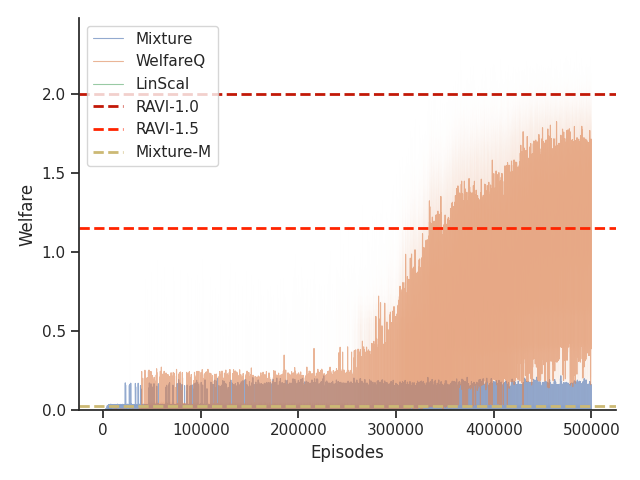}
        \caption*{$d=5$}
    \end{subfigure}%
    \caption{Taxi, $W_{p=0.001}$}
    \label{appendix:learning curve taxi p=0.001}
\end{figure*}

\begin{figure*}[h]
    \begin{subfigure}{0.25\linewidth}
        \centering
        \includegraphics[width=\linewidth]{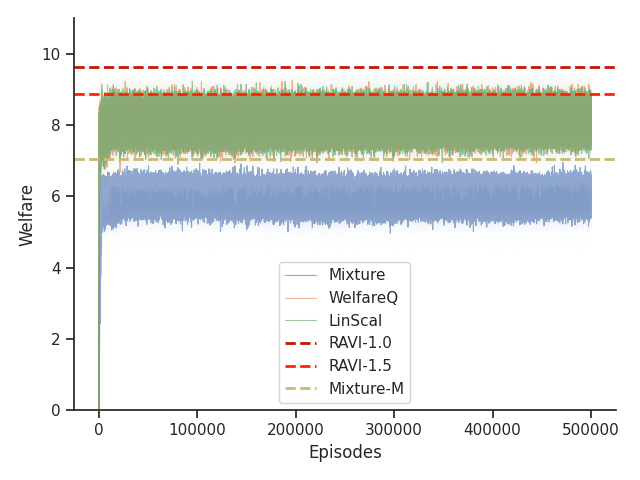}
        \caption*{$d=2$}
    \end{subfigure}%
    \begin{subfigure}{0.25\linewidth}
        \centering
        \includegraphics[width=\linewidth]{figs/taxi-p-welfare-0.9-15-3.png}
        \caption*{$d=3$}
    \end{subfigure}%
    \begin{subfigure}{0.25\linewidth}
        \centering
        \includegraphics[width=\linewidth]{figs/taxi-p-welfare-0.9-15-4.png}
        \caption*{$d=4$}
    \end{subfigure}%
    \begin{subfigure}{0.25\linewidth}
        \centering
        \includegraphics[width=\linewidth]{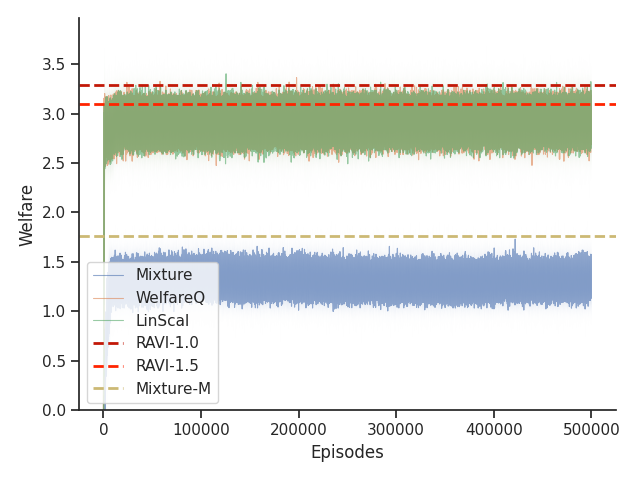}
        \caption*{$d=5$}
    \end{subfigure}%
    \caption{Taxi, $W_{p=0.9}$}
    \label{appendix:learning curve taxi p=0.9}
\end{figure*}

\begin{figure*}[h]
    \begin{subfigure}{0.33\linewidth}
        \centering
        \includegraphics[width=\linewidth]{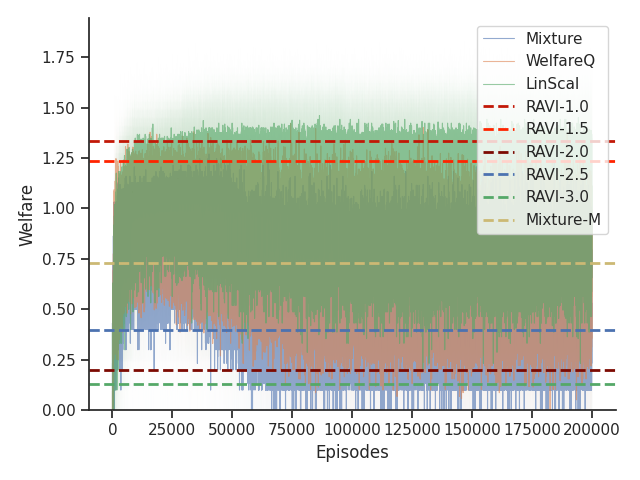}
        \caption*{$CD_{\rho=0.4}$}
    \end{subfigure}%
    \begin{subfigure}{0.33\linewidth}
        \centering
        \includegraphics[width=\linewidth]{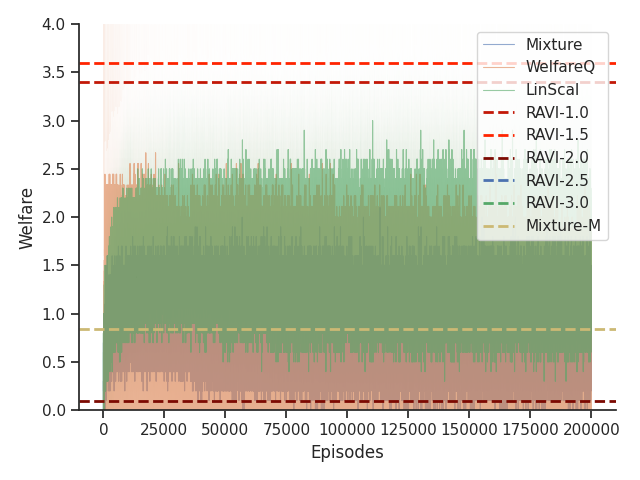}
        \caption*{$RD_{\text{threshold}=2}$}
    \end{subfigure}%
    \caption{Scavenger}
    \label{appendix:learning curve scavenger}
\end{figure*}

Given that \Cref{result:tab1} in the main text solely contains results evaluated from algorithms trained until convergence, it misses other important information such as the rate of convergence and the learning process of some online algorithms such as WelfareQ. Thus, in this subsection, we provide our visualizations of the learning curve for these algorithms. The plot is created using mean over 10 random initializations with standard deviation as shaded regions. Note that we use horizontal lines for model-based approaches as they do not have an online learning phase. Our results can be found at \Cref{appendix:learning curve taxi nash}, \Cref{appendix:learning curve taxi egalitarian}, \Cref{appendix:learning curve taxi p=-10}, \Cref{appendix:learning curve taxi p=0.001}, \Cref{appendix:learning curve taxi p=0.9}, and \Cref{appendix:learning curve scavenger}.


\FloatBarrier
\clearpage
\section{Details on the \textsc{RAEE} Algorithm}

In this section, we give a detailed description of the \textsc{RAEE} algorithm, extending from \cite{kearns02} and demonstrating Theorem~\ref{theorem:RA-E3}, the main theorem for the \textsc{RAEE} algorithm.

\subsection{Introducing \textsc{RAEE}}

The extension works as follows. The algorithm starts off by doing \textit{balanced wandering} \cite{kearns02}: when encountering a new state, the algorithm selects a random action. However, when revisiting a previously visited state, it chooses the least attempted action from that state, resolving ties by random action selection. At each state-action pair $(s, a)$ it tries, the algorithm stores the reward $\mathbf{R}(s, a)$ received and an estimate of the transition probabilities $Pr(s' \mid s, a)$ derived from the empirical distribution of next states reached during balanced wandering.

Next, we introduce the notion of a \textit{known state} \cite{kearns02}, which refers to a state that the algorithm has explored to the extent that the estimated transition probabilities for any action from that state closely approximate their actual values. Denote the number of times a state needs to be visited as $m_{known}$. We will specify the value later in our runtime characterization.

States are thus categorized into three groups: known states, which the algorithm has extensively visited and obtained reliable transition statistics; states that have been visited before but remain unknown due to limited trials and therefore unreliable data; and states that have not been explored at all. By the Pigeonhole Principle, accurate statistics will eventually accumulate in some states over time, leading to their becoming known. Let $S$ be the set of currently known states, the algorithm can build the current known-state MOMDP $M_S$ that is naturally induced on $S$ by the full MOMDP $M$ with all ``unknown'' states merged into a single absorbing state. Although the algorithm cannot access $M_S$ directly, it will have an approximation $\hat{M}_S$ by the definition of the known states. By the simulation lemma \cite{kearns02}, $\hat{M}_S$ will be an accurate model in the sense that the expected $T$-step welfare of any policy in $\hat{M}_S$ is close to its expected $T$-step return in $M_S$. (Here $T$ is the horizon time.) Hence, at any timestep, $\hat{M}_S$ functions as an incomplete representation of $M$, specifically focusing on the part of $M$ that the algorithm possesses a strong understanding of.

This is where we insert \textsc{RAVI}. The algorithm performs the two off-line optimal policy computations; i) first on $\hat{M}_S$ using \textsc{RAVI} to compute an \textit{exploitation} policy that yields an approximately optimal welfare and ii) second performing traditional value iteration on $\hat{M}'_S$, which has the same transition probabilities as $\hat{M}_S$, but different payoffs: in $\hat{M}'_S$, the absorbing state (representing ``unknown'' states) has scalar reward $1$ and all other states have scalar reward $0$. The optimal policy in $\hat{M}'_S$ simply exits the known model as rapidly as possible, rewarding exploration. 

By the explore or exploit lemma \cite{kearns02}, the algorithm is guaranteed to either output a policy with approximately optimal return in $M$, or to improve the statistics at an unknown state. Again by the Pigeonhole Principle, a new state becomes known after the latter case occurs for some finite number of times, and thus the algorithm is always making progress. In the worst case, the algorithm builds a model of the entire MOMDP $M$. Having described the elements, we now outline the entire extended algorithm, where the notations are consistent with \cite{kearns02}.
\\\\ \textbf{\textsc{RAEE} Algorithm:}
\begin{itemize}
    \item (Initialization) Initially, the set $S$ of known states is empty.
    \item (Balanced Wandering) Any time the current state is not in $S$, the algorithm performs balanced wandering
    \item (Discovery of New Known States) Any time a state $s$ has been visited $m_{known}$ \cite{kearns02} times during balanced wandering, it enters the known set $S$, and no longer participates in balanced wandering.
    \item (Off-line Optimizations) Upon reaching a known state $s \in S$ during balanced wandering, the algorithm performs the two off-line optimal policy computations on $\hat{M}_S$ and $\hat{M}'_S$ described above:
    \begin{itemize}
        \item (Attempted Exploitation) Use \textsc{RAVI} algorithm to compute an $\epsilon/2$-optimal policy on $\hat{M}_S$. If the resulting exploitation policy $\hat{\pi}$ achieves return from $s$ in $\hat{M}_S$ that is at least $V^*(s, \mathbf{0}, T) - \epsilon/2$, the algorithm halts and outputs $\hat{\pi}$.
        \item (Attempted Exploration) Otherwise, the algorithm executes the resulting exploration policy derived from the off-line computation on $\hat{M}'_S$ for $T$ steps in $M$.
    \end{itemize}
    \item (Balanced Wandering) Any time an attempted exploitation or attempted exploration visits a state not in $S$, the algorithm resumes balanced wandering.
\end{itemize}

This concludes the description of the algorithm.

\subsection{Runtime Analysis}

In this subsection, we comment on additional details of the analysis of Theorem~\ref{theorem:RA-E3}, the main theorem for the \textsc{RAEE} algorithm.\\

\noindent \textbf{Theorem~\ref{theorem:RA-E3}} (\textsc{RAEE}).
Let $V^*(s, 0, T)$ denote the value function for the policy with the optimal expected welfare in the MOMDP $M$ starting at state $s$, with $\mathbf{0} \in \reals^d$ accumulated reward and $T$ timesteps remaining. Then for a uniformly continuous welfare function $W$, there exists an algorithm $A$, taking inputs $\epsilon$, $\beta$, $|\mathcal{S}|$, $|\mathcal{A}|$, and $V^*(s, \mathbf{0}, T)$, such that the total number of actions and computation time taken by $A$ is polynomial in $1/\epsilon$, $1/\beta$, $|\mathcal{S}|$, $|\mathcal{A}|$, the horizon time $T = 1/(1 - \gamma)$ and exponential in the number of objectives $d$, and with probability at least $1 - \beta$, $A$ will halt in a state $s$, and output a policy $\hat{\pi}$, such that $V^{\hat{\pi}}_M (s, 0, T) \ge V^*(s, 0, T) -\epsilon$.\\

We begin by defining approximation of MOMDPs. Think of $M$ as the true MOMDP, that is, a perfect model of transition probabilities. Think of $M'$, on the other hand, as the current best estimate of an MOMDP obtained through exploration. In particular, $M'$ in the \textsc{RAEE} algorithm will consist of the set of known states.

\begin{definition}\cite{kearns02}
Let $M$ and $M'$ be two MOMDPs over the same state space $\mathcal{S}$ with the same deterministic reward function $\mathbf{R}(s, a)$. $M'$ is an $\alpha-$approximation of $M$ if for any state $s$ and $s'$ and any action $a$,
$$
Pr_M(s' \mid s, a) - \alpha \le Pr_{M'}(s' \mid s, a) \le Pr_M(s' \mid s, a) + \alpha,
$$
where the subscript denotes the model.
\end{definition}

We now extend the Simulation Lemma \cite{kearns02}, which tells us how close the approximation of an MOMDP needs to be in order for the expected welfare, or ESR, of a policy to be close in an estimated model. The argument is similar to \cite{kearns02}, but in our case the policy may not be stationary and we want to bound the deviation in expected welfare rather than just accumulated reward. 

Recall that $G^T_{max}$ is defined to be the maximum possible welfare achieved on a $T$-step trajectory - $G^T_{max}$ is at most $W(\mathbf{T})$ where $\mathbf{T}$ equals $T$ times the identity vector, in our model.

\begin{lemma}[Extended Simulation Lemma]
\label{lemma:simulation}
    Let $M'$ be an \\$O((\epsilon/|\mathcal{S}||\mathcal{A}| T G^T_{\max})^2)$-approximation of $M$. Then for any policy $\pi$, any state $s$, and horizon time $T$, we have
    $$
    V^\pi_M(s, \tau_{0;0}, T) - \epsilon \le V^\pi_{M'}(s, \tau_{0;0}, T) \le V^\pi_M(s, \tau_{0;0}, T) + \epsilon.
    $$
\end{lemma}

\begin{proof}
    Fix a policy $\pi$ and a start state $s$. Let $M'$ be an $\alpha$-approximation of $M$ (we will later show that $\alpha$ has the same bound as the Lemma statement). Call the transition probability from a state $s$ to a state $s'$ under action $a$ to be $\beta$-small in $M$ if $Pr_M(s' \mid s, a) \le \beta$. Then the probability that a $T$-trajectory $\tau$ starting from a state $s$ following policy $\pi$ contains at least one $\beta$-small transition is at most $\beta |\mathcal{S}| |\mathcal{A}| T$. This is because the total probability of all $\beta$-small transitions in $M$ is at most $\beta |\mathcal{S}| |\mathcal{A}|$ (assuming all transition probabilities are $\beta$-small), and there are $T$ timesteps. Note that in our case, the optimal policy may not be necessarily stationary, thus the agent does not necessarily choose the same action (and hence the same transition probability) upon revisiting any state. So we cannot bound the total probability by $\beta |\mathcal{S}|$ like in the original proof.
    
    The total expected welfare of the trajectories of $\pi$ that consist of at least one $\beta$-small transition of $M$ is at most $\beta |\mathcal{S}| |\mathcal{A}| T G^T_{\max}$. Recall that $M'$ be an $\alpha$-approximation of $M$. Then for any $\beta$-small transition in $M$, we have $Pr_{M'}(s' \mid s, a) \le \alpha + \beta$. So the total welfare of the trajectories of $\pi$ that consist of at least one $\beta$-small transition of $M$ is at most $(\alpha + \beta)|\mathcal{S}| |\mathcal{A}| T G^T_{\max}$. We can thus bound the difference between $V^\pi_M (s, \tau_{0;0}, T)$ and $V^\pi_{M'} (s, \tau_{0;0}, T)$ restricted to these trajetories by $(\alpha + 2\beta) |\mathcal{S}| |\mathcal{A}| T G^T_{\max}$. We will later choose $\beta$ and bound this value by $\epsilon/4$ to solve for $\alpha$.

    Next, consider trajectories of length $T$ starting at $s$ that do not contain any $\beta$-small transitions, i.e. $Pr_M(s' \mid s, a) > \beta$ in these trajectories. Choose $\Delta = \alpha / \beta$, we may write
    \begin{align*}
        &(1 - \Delta)Pr_M(s' \mid s, a) \\
        &\qquad \le Pr_{M'}(s' \mid s, a) \\
        &\qquad \le (1 + \Delta) Pr_M (s' \mid s, a).
    \end{align*}
    because $M'$ is an $\alpha$-approximation of $M$ and $\beta \leq 1$. Thus for any $T$-trajectory $\tau$ that does not cross any $\beta$-small transitions under $\pi$, we have
    $$
    (1 - \Delta)^TPr_M^\pi[\tau] \le Pr_{M'}^\pi[\tau] \le (1 + \Delta)^T Pr_M ^\pi[\tau],
    $$
    which follows from the definition of the probability along a trajectory and the fact that $\pi$ is the same policy in all terms. Since we assume reward functions are deterministic in our case, for any particular $T$-trajectory $\tau$, we also have
    $$
    W_M(\mathbf{R}(\tau))= W_{M'}(\mathbf{R}(\tau)).
    $$
    Since these hold for any fixed $T$-trajectory that does not traverse any $\beta$-small transitions in $M$ under $\pi$, they also hold when we take expectations over the distributions over such $T$-trajectories in $M$ and $M'$ induced by $\pi$. Thus
    \begin{align*}
        &(1 - \Delta)^TV^\pi_{M}(s, \tau_{0;0}, T) - \frac{\epsilon}{4} \\
        &\qquad \le V^\pi_{M'}(s, \tau_{0;0}, T) \\ 
        &\qquad \le (1 + \Delta)^T V^\pi_{M}(s, \tau_{0;0}, T) + \frac{\epsilon}{4}.    
    \end{align*}
    where the $\epsilon/4$ terms account for the contributions of the $T$ -trajectories that traverse at least one $\beta$-small transitions under $\pi$. It remains to show how to choose $\alpha$, $\beta$, and $\Delta$ to obtain the desired approximation. For the upper bound, we solve for
    \begin{align*}
        (1 + \Delta)^T V^\pi_{M}(s, \tau_{0;0}, T) \le V^\pi_{M}(s, \tau_{0;0}, T) +  \frac{\epsilon}{4} \\
        \implies (1+\Delta)^T \le 1 + \epsilon/(4G^T_{\max}).   
    \end{align*}
    By taking $\log$ on both sides and using Taylor expansion, we can upper bound $\Delta$.
    \begin{align*}
        &T\Delta/2 \le \epsilon/(4G^T_{\max}) \\
        &\qquad \implies \Delta \le \epsilon/(2TG^T_{\max}).
    \end{align*}
    Choose $\beta = \sqrt{\alpha}$. Then
    $$
    \begin{cases}
        (\alpha + 2\beta) |\mathcal{S}| |\mathcal{A}| T G^T_{\max} \le 3\sqrt{\alpha} |\mathcal{S}| |\mathcal{A}| T G^T_{\max} \le \epsilon/4 \\
        \Delta = \sqrt{\alpha} \le \epsilon/(2TG^T_{\max})
    \end{cases}
    $$
    Choosing $\alpha = O((\epsilon/|\mathcal{S}||\mathcal{A}| T G^T_{\max})^2)$ solves the system. The lower bound can be handled similarly, which completes the proof of the lemma.
\end{proof}

We now define a ``known state.'' This is a state that has been visited enough times, and its actions have been trialed sufficiently many times, that we have accurate estimates of the transition probabilities from this state.

\begin{definition}\cite{kearns02}
    Let $M$ be an MOMDP. A state $s$ of $M$ is considered \textit{known} if it has be visited a number of times equal to
    $$
    m_{known} = O((|\mathcal{S}||\mathcal{A}| T G^T_{\max}/\epsilon)^4 |\mathcal{A}| \log(1/\delta)).
    $$
\end{definition}

By applying Chernoff bounds, we can show that if a state has been visited $m_{known}$ times then its empircal estimation of the transition probabilities satisfies the accuracy required by the Lemma~\ref{lemma:simulation}.

\begin{lemma}
\label{lemma:known}
\cite{kearns02}
    Let $M$ be an MOMDP. Let $s$ be a state of $M$ that has been visited at least $m$ times, with each action having been executed at least $\lfloor m/|\mathcal{A}| \rfloor$ times. Let $\hat{Pr}(s' \mid s ,a)$ denote the empirical transition probability estimates obtained from the $m$ visits to $s$. Then if
    $$
    m = O((|\mathcal{S}||\mathcal{A}| T G^T_{\max}/\epsilon)^4 |\mathcal{A}| \log(1/\delta)),
    $$
    then with probability $1-\delta$, we have
    $$
    |\hat{Pr}(s' \mid s ,a) - Pr(s' \mid s, a)| = O((\epsilon/|\mathcal{S}||\mathcal{A}| T G^T_{\max})^2)
    $$
    for all $s' \in \mathcal{S}$.
\end{lemma}

\begin{proof}
    The sampling version of Chernoff bounds states that if the number of independent, uniformly random samples $n$ that we use to estimate the fraction of a population with certain property $p$ satisfies
    $$
    n = O\left( \frac1{\alpha^2} \log\left(\frac1\delta\right) \right),
    $$
    the our estimate $\bar{X}$ satisfies
    $$
    \bar{X} \in [p - \alpha, p + \alpha] ~\text{with probability $1-\delta$}.
    $$
    By the Extended Simulation Lemma, it suffices to choose $\alpha = O((\epsilon/|\mathcal{S}||\mathcal{A}| T G^T_{\max})^2)$.
    
     Note that we need to insert an extra factor of $|\mathcal{A}|$ compared to the original analysis since we treat the size of the action space as a variable instead of a constant, and a state is categorized as ``known'' only if the estimates of transition probability of \textit{all} actions are close enough.
\end{proof}

We have specified the degree of approximation required for sufficient simulation accuracy. It remains to directly apply the Explore or Exploit Lemma from \cite{kearns02} to conclude the analysis.

\begin{lemma}[Explore or Exploit Lemma \cite{kearns02}]
    Let $M$ be any MOMDP, let $S$ be any subset of the states of $M$, and let $M_S$ be the MOMDP on $M$. For any $s \in \mathcal{S}$, any $T$, and any $1 > \alpha > 0$, either there exists a policy $\pi$ in $M_S$ such that $V^\pi_{M_S}(s, \tau_{0;0}, T ) \ge V^*_M(s, \mathbf{0}, T) - \alpha$, or there exists a policy $\pi$ in $M_S$ such that the probability that a $T$-trajectory following $\pi$ will lead to the exit state exceeds $\alpha / G^T_{\max}$.
\end{lemma}

This lemma guarantees that either the $T$ -step return of the \textit{optimal} exploitation policy in the simulated model is very close to the optimal achievable in $M$, or the agent choosing the exploration policy can reach a previously unknown state with significant probability. 

Note that unlike the original E3 algorithm, which uses standard value iteration to compute the exactly optimal policy (optimizing a linear function of a scalar reward) on the sub-model $\hat{M}_S$, we use our \textsc{RAVI} algorithm to find an \textit{approximately} optimal policy. Therefore, we need to allocate $\epsilon/2$ error to both the simulation stage \textit{and} the exploitation stage, which gives a total of $\epsilon$ error for the entire algorithm.

It remains to handle the failure parameter $\beta$ in the statement of the main theorem, which can be done similarly to \cite{kearns02}. There are two sources of failure for the algorithm:
\begin{itemize}
    \item The algorithm's estimation of the next state distribution for some action at a known state is inaccurate, resulting $\hat{M}_S$ being an inaccurate model of $M_S$.
    \item Despite doing attempted explorations repeatedly, the algorithm fails to turn a previously unknown state into a known state because of an insufficient number of balanced wandering.
\end{itemize}
It suffices to allocate $\beta/2$ probabilities to each source of failure. The first source of failure are bounded by Lemma \ref{lemma:known}. By choosing $\beta' = \beta/(2|\mathcal{S}|)$, we ensure that the probability that the first source of failure happens to each of the known states in $M_S$ is sufficiently small such that the total failure probability is bounded by $\beta/2$ for $M_S$.

For the second source of failure, by the Explore or Exploit Lemma, each attempted exploration results in at least one
step of balanced wandering with probability at least $\epsilon/(2G^T_{\max})$, i.e. when this leads the agent to an unknown state. The agent does at most $|\mathcal{S}| m_{known}$ steps of balanced wandering, since this makes every state known. By Chernoff bound, the probability that the agent does fewer than $|\mathcal{S}| m_{known}$ steps of balanced wandering (attempted explorations that actually leads to an unknown state) will be smaller than $\beta/2$ if the number of attempted explorations is
$$
O((G^T_{\max}/\epsilon)|\mathcal{S}| \log(|\mathcal{S}|/\beta)m_{known}),
$$
where $m_{known} = O(((|\mathcal{S}||\mathcal{A}|TG^T_{\max})/\epsilon)^4 |\mathcal{A}|\log(|\mathcal{S}|/\beta))$ (recall we choose $\beta' = \beta/(2|\mathcal{S}|)$ for the first source of failure).

Thus, the total computation time is bounded by $O\left(|\mathcal{S}|^2 |\mathcal{A}| (T/\alpha_{T\epsilon/2})^d \right)$ (the time required for \textsc{RAVI} during off-line computations with $\epsilon/2$ precision by Lemma \ref{lemma:error}) times $T$ times the maximum number of attempted explorations, giving
$$
O((|\mathcal{S}|^3 |\mathcal{A}| T G^T_{\max}/\epsilon) (T/\alpha_{T\epsilon/2})^d \log(|\mathcal{S}|/\beta)m_{known}).
$$

This concludes the proof of Theorem~\ref{theorem:RA-E3}.

\section{More Discussion about experiments \label{appendix:exp discussion}}
\textbf{Deterministic Transitions:}  Deterministic settings are common in many state-of-the-art environments in MORL, particularly when focusing on episodic tasks with short time horizons. These settings are often used to emphasize the core algorithmic contributions without introducing additional complexities from stochastic transitions or long horizons. Nonetheless, we acknowledge that incorporating stochastic environments would better showcase the generality of our approach and highlight this as an important direction for future work.

\textbf{Model Sizes and Scalability:} We recognize that the model sizes used in the experiments are relatively small, primarily due to memory limitations in the tabular setting, and that function approximation would be needed to move to large state space environments as highlighted in our future work.

\textbf{RAEE Algorithm:} RAVI is a model-based approach, and one of the baselines, ``Mixture-M'' for a model-based mixture policy, uses the model. For the others model-free baselines, we only compare performance at convergence with RAVI to achieve a more fair comparison. To the best of our knowledge, there are no other model-based algorithms in the field specifically designed for optimizing ESR objectives. Due to computational and space constraints, we focused our empirical evaluation on RAVI, as the optimization subroutine is the core algorithmic contribution. That said, we acknowledge this limitation and believe benchmarking RAEE empirically is a worthwhile direction for future research.

\textbf{Model Accuracy and RAEE:} While it is true that RAVI’s performance depends on model accuracy, the central goal of RAEE is precisely to ensure it learns a sufficiently accurate model of the environment to provably guarantee an approximation factor. For future work, comparing RAVI with RAEE would provide valuable insights into the trade-offs between leveraging accurate models and learning them through exploration.

\end{document}